\documentclass[lettersize,journal]{IEEEtran}
\usepackage{amsmath,amsfonts}
\usepackage{algorithmic}
\usepackage{algorithm}
\usepackage{array}
\usepackage[caption=false,font=normalsize,labelfont=sf,textfont=sf]{subfig}
\usepackage{textcomp}
\usepackage{stfloats}
\usepackage{url}
\usepackage{verbatim}
\usepackage{graphicx}
\usepackage{booktabs}
\usepackage{multirow}
\usepackage[table]{xcolor}

\usepackage{pifont}
\usepackage{bm}
\usepackage{amssymb}
\usepackage{cleveref}
\usepackage{subcaption}
\usepackage{cite}
\begin{document}

\title{Geometric Constrained Non-Line-of-Sight Imaging}

\author{Xueying Liu, Lianfang Wang, Jun Liu, Yong Wang
        and Yuping Duan

\IEEEcompsocitemizethanks{
\IEEEcompsocthanksitem X. Liu is with the Center for Applied Mathematics, Tianjin University, Tianjin 300072, China.
\IEEEcompsocthanksitem Y. Wang is with the School of Physics, Nankai University, China.
\IEEEcompsocthanksitem L. Wang, J. Liu and Y. Duan are with the School of Mathematical Sciences, Beijing Normal University, Beijing, 100875, China.
E-mail: doveduan@gmail.com}
\thanks{This paper was produced by the IEEE Publication Technology Group. They are in Piscataway, NJ.}
\thanks{Manuscript received April 19, 2021; revised August 16, 2021.}}

\markboth{Journal of \LaTeX\ Class Files,~Vol.~14, No.~8, August~2021}%
{Shell \MakeLowercase{\textit{et al.}}: A Sample Article Using IEEEtran.cls for IEEE Journals}


\maketitle

\begin{abstract}
Normal reconstruction is crucial in non-line-of-sight (NLOS) imaging, as it provides key geometric and lighting information about hidden objects, which significantly improves reconstruction accuracy and scene understanding. However, jointly estimating normals and albedo expands the problem from matrix-valued functions to tensor-valued functions that substantially increasing complexity and computational difficulty. In this paper, we propose a novel joint albedo-surface reconstruction method, which utilizes the Frobenius norm of the shape operator to control the variation rate of the normal field. It is the first attempt to apply regularization methods to the reconstruction of surface normals for hidden objects. By improving the accuracy of the normal field, it enhances detail representation and achieves high-precision reconstruction of hidden object geometry. The proposed method demonstrates robustness and effectiveness on both synthetic and experimental datasets. On transient data captured within 15 seconds, our surface normal-regularized reconstruction model produces more accurate surfaces than recently proposed methods and is 30 times faster than the existing surface reconstruction approach.
\end{abstract}

\begin{IEEEkeywords}
Non-line-of-sight imaging, under-sampled scanning, surface normal, shape operator, surface regularization.
\end{IEEEkeywords}
\begin{figure*}[h]
      \begin{center}			\includegraphics[width=0.9\linewidth]{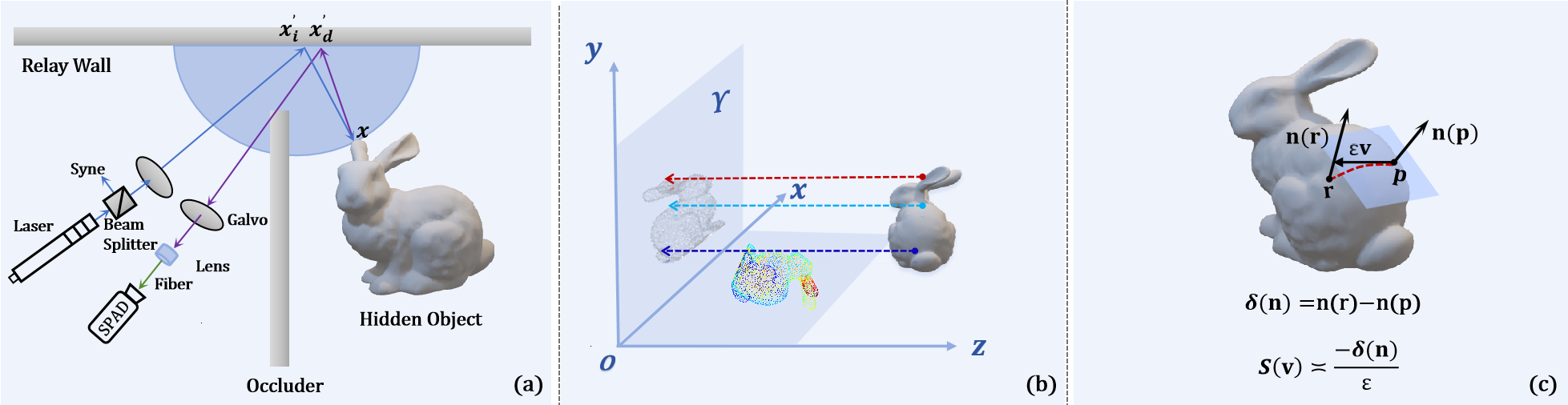}
	  \end{center}
	\caption{Illustration of our geometric constrained non-line-of-sight imaging framework, where (a) NLOS system, (b) projection of volumetric albedo to albedo image (left side) and depth image (bottom), and (c) illustration of shape operator on surface.}
	\label{illustration}
\end{figure*}
\section{Introduction}
\IEEEPARstart{I}{n} typical Non-Line-of-Sight (NLOS) imaging, a hidden target is detected through light scattered off a surface, which, while unclear, can now reveal hidden objects thanks to advances in sensitive optics and reconstruction methods \cite{faccio2020non}. Using time-resolved measurements, NLOS imaging recovers the 3D shape and appearance of objects outside the sensor’s line of sight, with applications in autonomous driving \cite{scheiner2020seeing}, 3D human pose estimation \cite{Isogawa0OK20}, and sensor systems \cite{chen2020learned}, among others \cite{bertolotti2012non, katz2014non, saunders2020multi}.

NLOS imaging can be categorized into two types: active \cite{nam2021low,pei2021dynamic,cao2022high} and passive \cite{batarseh2018passive,geng2021passive}. Passive NLOS imaging relies on ambient light, often resulting in high noise and low resolution. In contrast, active NLOS imaging uses photons from an active light source, capturing transient signals via time-of-flight (TOF) measurements \cite{kadambi2013coded}, enabling effective spatiotemporal data collection. Common hardware setups include streak cameras with pulsed lasers, single-photon avalanche diodes (SPADs) \cite{buttafava2015non} with pulsed lasers, and AMCW ToF cameras with modulated light sources, with SPADs and pulsed lasers being the most widely used. We focus on transient-based NLOS imaging, where hidden scene representation methods are divided into two categories: volumetric albedo reconstruction \cite{arellano2017fast,velten2012recovering,o2018confocal,liu2020phasor,lindell2019wave} and object surface reconstruction \cite{tsai2017geometry,heide2019non,xin2019theory,tsai2019beyond,young2020non,fujimura2023nlos}. Volumetric methods estimate albedo values of 3D voxels using techniques like filtered back-projection \cite{velten2012recovering,la2018error,arellano2017fast}, light cone transform (LCT) \cite{o2018confocal}, and phasor field (PF) \cite{liu2020phasor}. Surface-based methods recover object surfaces by estimating surface normals, offering potential for detailed geometry capture \cite{ju2023gr}. However, current techniques face challenges such as limited applicability to complex scenes \cite{tsai2017geometry}, sensitivity to noise \cite{young2020non} and initialization \cite{xin2019theory}, and high computational costs \cite{tsai2019beyond}.

For practical non-line-of-sight (NLOS) imaging, rapid data acquisition is essential. One approach is to improve the imaging system itself \cite{nam2021low,pei2021dynamic,callenberg2021low}, while a more direct strategy involves reducing illumination points and shortening exposure times. However, this can result in incomplete data and higher noise levels, limiting the ability of direct methods to produce high-quality reconstructions. Iterative reconstruction methods leverage prior knowledge about transient measurements and objects to enhance image quality \cite{heide2014diffuse,ahn2019convolutional,ye2021compressed,liu2021non,huang2023non,ding2024curvature,liu2023few,liu2024rm}. Techniques such as sparse priors, LASSO regularization \cite{ye2021compressed}, and total variation regularization \cite{liu2022photon} have been applied for target reconstruction. Additionally, collaborative regularization across multiple domains has been explored to restore low-quality transient signals and hidden details. For instance, methods like SOCR \cite{liu2021non} and CNLOS \cite{ding2024curvature}, combined with volume-surface collaborative regularization (SSCR) \cite{liu2023few}, have been proposed. However, SOCR and SSCR rely on complex regularization techniques, while CNLOS introduces high-order gradient terms into the 3D volume. Both approaches significantly increase computational complexity, potentially offsetting the time savings from reduced data acquisition. Recent advances in deep learning have enabled neural networks to address NLOS imaging challenges \cite{su2018deep,chen2020learned,li2023nlost,zhu2023compressive,liu2023non,ye2024plug,shen2021non}. Yet, these methods heavily depend on training datasets. The lack of real-world NLOS imaging datasets remains a limitation, and mismatches between the distribution of real-world data and training data may significantly degrade reconstruction performance.

In this work, we propose a joint reconstruction model for volumetric albedo and surface details tailored for under-sampled transient data. As illustrated in Figure \ref{illustration}(b), our approach approximates volumetric albedo using two-dimensional depth and albedo maps. By applying effective regularization techniques to these maps, we significantly enhance the reconstruction quality of hidden scenes. Unlike LCT, which ignores normal information, and DLCT, which directly estimates normal albedo, we introduce a geometric prior for surface reconstruction to characterize the rate of change in surface normals, specifically by regularizing the shape operator to constrain the gradient field of the depth map (as shown in Figure \ref{illustration}(c)). Our method not only improves reconstruction quality by mitigating blurring artifacts caused by the complexity of light propagation paths but also significantly reduces computational complexity by incorporating higher-order geometric constraints on the two-dimensional depth and albedo maps. To sum up, our main contributions are provided as follows
\begin{itemize}
\item We propose a joint reconstruction model for transient data, objects, and surfaces to address the challenges of under-sampled transient data under sparse illumination points and short exposure times.
\item We introduce an effective geometric constraint for target surface reconstruction, enforcing sparsity in the gradient field of normals by minimizing the Frobenius norm of the shape operator, thereby ensuring surface continuity.
\item We develop an efficient numerical algorithm that significantly reduces computational cost by implementing higher-order regularization and explicit computation strategies on two-dimensional depth and albedo functions.
\end{itemize}

\section{Related works}
\label{sec:related works}

Let $\bm{x}=(x,y,z)$ be the three-dimensional scene coordinates, and
$\bm{x}_i^{\prime}=(x_{i}^{\prime},y_{i}^{\prime},z = 0)$, $\bm{x}_d^{\prime}=(x_{d}^{\prime},y_{d}^{\prime},z = 0)$ be the illumination and detection coordinates on the visual wall, respectively.
The laser sends a pulse of light towards point $\bm{x}_i^{\prime}$ on a visible wall, and the light diffusely scatters from that point. The scattered light illuminates the objects hidden around a corner, and a fraction of that light reflects back towards the wall in response. A transient sensor measures the temporal response at a point $\bm{x}_d^{\prime}$ on the wall. The transient measurement, $\tau(\bm{x}_i^{\prime}, \bm{x}_d^{\prime}, t)$, represents the amount of light detected at point $\bm{x}_d^{\prime}$ at time $t$. For simplicity, we omit the travel time between the system and the wall, which can be adjusted based on the wall's geometry relative to the laser and sensor positions. The image formation model is then given by
\begin{equation}
\begin{aligned}\tau(\bm{x}_i^{\prime},\bm{x}_d^{\prime},t)=&\iiint_\Omega\frac{\left\langle {{\bm{x}_i^{\prime }} - {\bm{x}}},\bm{n}({\bm{x}}) \right\rangle }
{\left\|\bm{x}_i^{\prime}-\bm{x}\right\| ^3}\cdot\frac{\left\langle {{\bm{x}_d^{\prime }} - {\bm{x}}},\bm{n}({\bm{x}}) \right\rangle }{\left\|\bm{x}_d^{\prime}-\bm{x}\right\| ^3}\\&\cdot\delta\Big(\left\|\bm{x}-\bm{x}_i^{\prime}\right\|+\left\|\bm{x}-\bm{x}_d^{\prime}\right\|-tc\Big)u(\bm{x})d\bm{x}.\end{aligned}
\label{nonconfocal-FM}
\end{equation}
The function $u(\bm{x})$ represents the albedo of objects at each point $\bm{x}$ within the 3D half-space $\Omega$ where $z > 0$. Here, $c$ denotes the speed of light, and $\bm{n}(\bm{x}) = (n_x,n_y,n_z)(\bm{x})$ is the surface normal vector. The symbol $\delta$ refers to the point-spread function (PSF). The expression
inside the  $\delta$ 
relates the distance light travels through the hidden volume to its time of flight. The denominator accounts for the decrease in the intensity of light as a function of distance traveled. When the illumination point and detection point are located at the same position, i.e., $\bm{x}^{\prime}=\bm{x}_i^{\prime}=\bm{x}_d^{\prime}$, the model \eqref{nonconfocal-FM} reduces to
\begin{equation*}
\begin{aligned}\tau(\bm{x}^{\prime},t)=\iiint_\Omega&\frac{\left\langle {{\bm{x}^{\prime }} - {\bm{x}}},\bm{n}({\bm{x}}) \right\rangle }{\left\|\bm{x}^{\prime}-\bm{x}\right\| ^3}\cdot\frac{\left\langle {{\bm{x}^{\prime }} - {\bm{x}}},\bm{n}({\bm{x}}) \right\rangle }{\left\|\bm{x}^{\prime}-\bm{x}\right\| ^3}\\&\cdot\delta\Big(2\left\|\bm{x}-\bm{x}^{\prime}\right\|-tc\Big)u(\bm{x})d\bm{x}.\end{aligned}
\label{eqnlos2}
\end{equation*}

\subsection{The Light-cone transform (LCT)}
In \cite{o2018confocal}, by assuming that light scattering on the object surface is isotropic, it can further simplify the geometry of light propagation as follows
\begin{equation}
\begin{aligned}
\tau(\bm{x}^{\prime},t)=\iiint_\Omega\frac{u(\bm{x})}{\left\|\bm{x}^{\prime}-\bm{x}\right\|^4}\delta\Big(2\left\|\bm{x}^{\prime}-\bm{x}\right\|-tc\Big)d\bm{x},
\end{aligned}
\label{eqconnlos}
\end{equation}
the discrete problem of which gives
\begin{equation}
\tau=Au
\label{ip}
\end{equation}
with $A$ being the light transport matrix. 
The inverse problem \eqref{ip} can be efficiently solved via Wiener filtering in the Fourier domain, known as the light-cone transform (LCT). While LCT effectively reconstructs the shapes of hidden objects, as a low-pass filter, it fails to accurately model the light reflection behavior of surface materials with anisotropic scattering properties.

\subsection{The Directional Light-cone transform (DLCT)}
Young et al. \cite{young2020non} introduced the directional-albedo variable, i.e., $\bm v (v_x,v_y,v_z)(\bm x)= u(\bm x)\bm n(\bm x)$, allowing to obtain the complex surface details. Thus, the forward model \eqref{nonconfocal-FM} becomes
\begin{equation*}
\begin{aligned}\label{dirtional-model}
\tau ({{\bm{x}}^\prime },t) \!=\! \iiint_\Omega  {\frac{{\left\langle {\bm v({\bm{x}}),{{\bm{x}}^\prime } - {\bm{x}}} \right\rangle }}{{{{\left\| {{{\bm{x}}^\prime } - {\bm{x}}} \right\|}^5}}}} \cdot \delta (2\left\| {{\bm{x}} - {{\bm{x}}^\prime }} \right\| - tc)d{\bm{x}}.
\end{aligned}
\end{equation*}
The corresponding inverse problem is a vectorial form of the model \eqref{ip} and can be efficiently solved using Cholesky-Wiener decomposition. The directional LCT model simulates light reflection and propagation on surfaces with complex geometries and materials. While effective at detailing fine surface structures, it struggles to represent an object's overall shape and potentially overlook large-scale features or complex forms due to amplified noise. Additionally, since the light transport matrix
$A$ acts as a low-pass filter with a high condition number, both LCT and DLCT reconstructions can degrade significantly in the presence of data loss or noise.

\section{Our Method}
\subsection{Sparse NLOS}

NLOS imaging can be accelerated using transient data with short exposure times and sparse illumination points, which can be mathematically formulated as follows 
\begin{equation}
    \tau_0 = SAu + \epsilon,
\label{sparseprob}
\end{equation}
where \(\epsilon\) is random noise, \(S\) is the selection matrix, and \(\tau_0\) is transient data acquired using sparse illumination points. We aim to solve the inverse problem \eqref{sparseprob} to estimate the hidden object \(u\) from \(\tau_0\). Using Bayesian estimation, we combine observed data with prior knowledge to obtain the solution. Specifically, we integrate the data with the prior distribution to derive the posterior distribution, addressing uncertainty and noise in the data. The joint probability of observing both $u$ and $\tau_0$ is given as follows
\[ P(u|\tau_0) = \frac{P(\tau_0|u) P(u) }{P(\tau_0)},\]
where $P(u|\tau_0)$ is the posterior probability, $P(\tau_0|u)$ is the likelihood function, $P(u)$ is the prior probability and $P(\tau_0)$ is the marginal likelihood. By introducing $\tau$, we can reformulate it as
\[P(u,\tau|\tau_0) =\frac{P(\tau_0|u,\tau)P(\tau|u) P(u)}{P(\tau_0)}.\]
We assume that \(u\) depends solely on the reconstructed transient data, i.e., $ P(\tau_0 | u, \tau) = P(\tau_0 | \tau)$. Based on Bayes' theorem and the Maximum a Posteriori (MAP) estimator, maximizing the posterior probability of $u$ and $\tau$ given $\tau_0$ is equivalent to minimizing the negative log-likelihood
\begin{equation} \label{map-u}
    (u^*,\tau^*) = \arg\min_{u,\tau} -\log P(\tau_0|\tau) - \log P(\tau|u) - \log P(u).
\end{equation}
Assuming the noise in the transient data is additive Gaussian noise and incorporating the sparsity prior of the 3D object, we derive the minimization problem from \eqref{map-u} as
\begin{equation}\label{object-recon}
    E_1(u,\tau)\!=\!\frac{1}{2}\int_\Omega(Au-\tau)^2 +\frac{\rho}{2}\int_{\Omega}(S\tau-\tau_0)^2+\gamma\int_\Omega|u|dx,
\end{equation}
where $\rho$ governs the consistency between the updated and initial signals, and $\gamma$ adjusts the weight of the sparsity term, defined as
\[\gamma = \sigma* \sum_{\substack{x, y, t}} \tau_0(x, y, t)\]
with $\sigma$ being a tunable positive parameter. Here, $\gamma$ is used to model the dependency relationship between  $u$ and $\tau_0$.

\subsection{Geometric constrained surface reconstruction}
Object reconstruction aims to align with the NLOS target surface where photons land. 
We project the surface onto a two-dimensional plane and utilize the depth map of the surface to represent the object's surface. As Fig. \ref{illustration} (b), let $\Upsilon:=(x,y,z=0)\in\Omega$ define the projection plane of the hidden object. The position of each point is determined by its distance to the projection plane.
For a given position $(x,y)\in\Upsilon$, let $n_{xy}$ denote the number of nonzero $u(x,y,:)$ values along the depth direction. If $n_{xy}$ is nonzero, the corresponding depth values are represented by $z_i$ for $i=1,\ldots,n_{xy}$, ordered from smallest to largest. A depth image $D$ and an albedo image $I$ can be derived from $u$ using the weighted projection function $(I,D):=P(u)$, defined as
\begin{equation*}
\begin{cases}
I_{xy}:=(P_I(u))_{xy}=\chi(n_{xy}) \cdot \sum_{i=1}^{n_{xy}} \omega_{xyz_i}\cdot u_{xyz_i}, \\
\\
D_{xy}:=(P_D(u))_{xy}=\Big[\chi(n_{xy}) \cdot \sum_{i=1}^{n_{xy}}\omega_{xyz_i} \cdot z_i\Big],
\end{cases}
\end{equation*}
where $\chi(\cdot)$ is the indicator function is $\chi(x) = 1$ for $x\neq 0$ and $\chi(x) = 0$ for $x =0$, and $[\cdot]$ is the rounding operation. The weights are estimated using
\begin{equation}\label{weight}
\omega_{xyz_{i}}=u_{xyz_{i}}^p/\sum_{j=1}^{n_{xy}}u_{xyz_{j}}^p,
\end{equation}
where $p$ denotes the power of $u$. It indicates that higher albedo contributes greater weight.

\begin{figure}[t]
      \begin{center}			
      \includegraphics[width=0.8\linewidth]{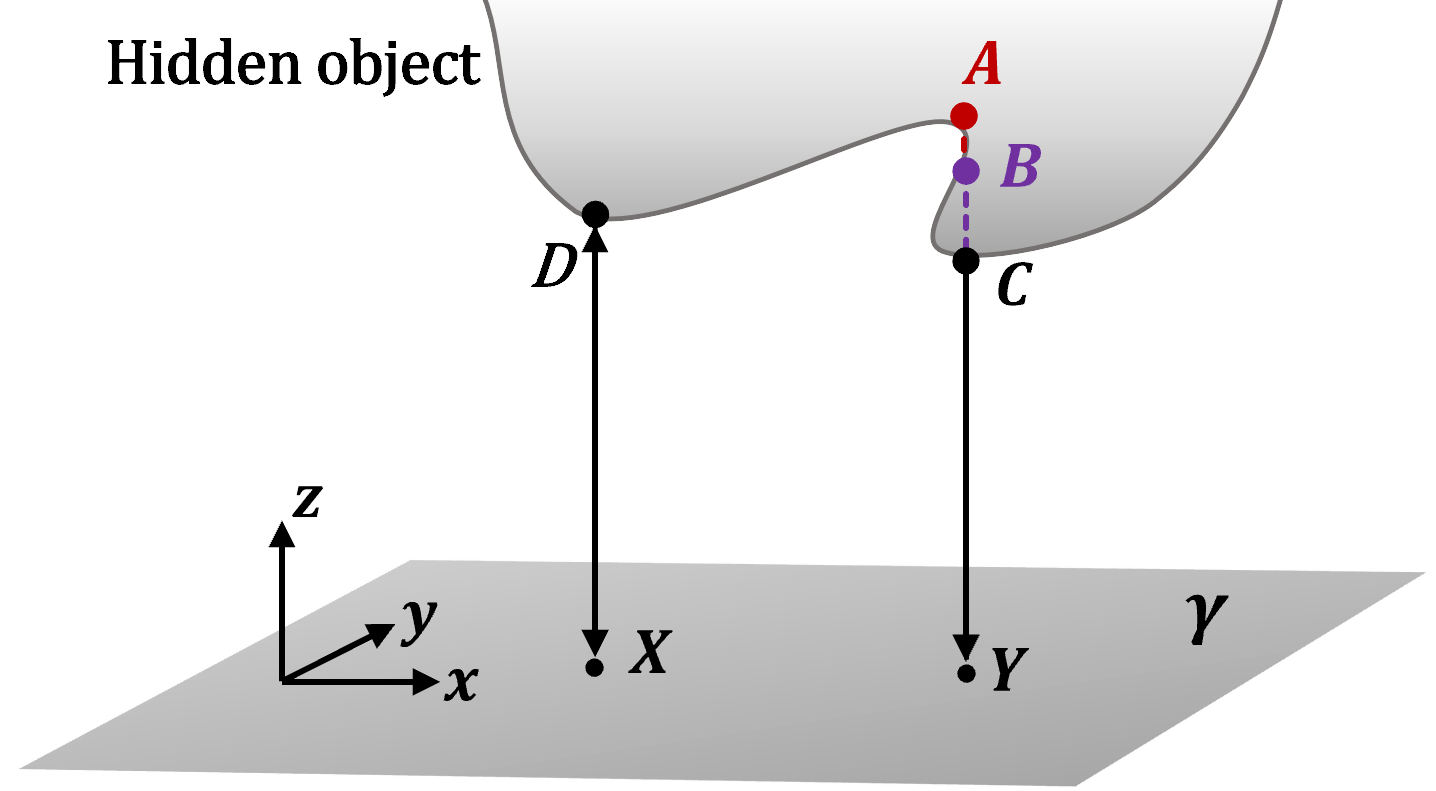}
	  \end{center}
	\caption{The projection operator between volumetric albedo and 2D albedo map. }
	\label{projection}
\end{figure}

\begin{algorithm*}[th]
\caption{\label{alg1}The ADMM-based algorithm for the joint reconstruction model}
\begin{algorithmic}[htbp]
\STATE \textbf{Input}: Raw data $\tau_0$, and parameters  $\gamma$, $\rho$, $\eta$, $\lambda$, $I^0=0$, $D^0=0$, ${K}_{max}$;
\STATE /* Initialize the variable $u$  */
\STATE $u^0=\arg\min\limits_u\frac12\int_\Omega(A_Su-\tau_0)^2dx+\gamma\int_\Omega|u|dx$;
\FOR{{$k=0$} \TO {$K_{max}-1$}}
  \STATE /* Compute $\tau^{k+1}$ */
    \STATE $\tau^{k+1}=\arg\min\limits_\tau~\frac12\int_\Omega(Au^{k}-\tau)^2dx+\frac{\rho}2\int_{\Omega}(S\tau-\tau_0)^2dx$;
    \STATE /* Compute $D^{k+1}$*/
    \STATE $D^{k+1} = \arg\min\limits_{D}~\frac12\int_{\Upsilon}(D-P_D(u^k))^2dx+ \int_{\Upsilon}|\alpha(D^k)||\nabla D|dx+\int_\Upsilon\beta(D^k)|\nabla^2D|_Fdx$;
    \STATE /* Compute $I^{k+1}$ */
    \STATE $I^{k+1} = \arg\min\limits_{I}~\frac12\int_{\Upsilon}(I-P_I(u^k))^2dx+\eta\int_{\Upsilon}|\nabla I|dx$;
    \STATE /* Compute $u^{k+1}$ */
    \STATE $u^{k+1}=\arg\min\limits_u~\frac12\int_\Omega(Au-\tau^{k+1})^2dx+\frac{\lambda}2\int_\Omega(u-P^{\dagger}(I^{k+1},D^{k+1}))^2dx+\gamma\int_\Omega|u|dx$;
\ENDFOR
\STATE \textbf{Output}: $u^{K_{max}}$.
\end{algorithmic}
\end{algorithm*}

As shown in Fig. \ref{projection}, a non-reversible many-to-one mapping is established, particularly for non-convex surfaces. However, the hidden object $u$ can be approximated using a depth image
$D$ and an albedo image
$I$ via back-projection
\begin{equation*}
 u_{xyD_{xy}}:=P^{\dagger}(I,D)=\begin{cases}
I_{xy}, \qquad \text{for} ~~ D_{xy}\neq 0,\\
0, \qquad~~\, \text{for} ~~ D_{xy}=0,
\end{cases}
\end{equation*}
where $P^{\dagger}(\cdot,\cdot)$ is the selective inverse function used to reconstruct the hidden object.
Let $\mathbb S=(x,y,D(x,y))$ be the surface defined in $(x,y)\in\Upsilon$, which corresponds to the zero level set of $\phi(x,y,z) = z-D(x,y)$. Since surface normals are crucial for accurately estimating and recovering the shape of the surface, we first define the \emph{unit normal} of the zero level set $\{(x,y,z):\phi(x,y,z)=0\}$ as \cite{osher2004level}
\[N_D=\frac{\nabla \phi}{|\nabla \phi|} = \frac{(\nabla D,-1)}{\sqrt{1+|\nabla D|^2}},\]
where $\nabla D$ is the gradient of $D$.
Surface normal smoothness is a key prior to reduce noise and unnatural discontinuities. We use the shape operator as a regularization for the depth map, as it captures the surface geometry, including curvature and bending. It provides a linear map that relates changes in the normal vector along the surface. Specifically, the shape operator is defined as:
\begin{equation*}
\begin{split}
    \mathcal S(D)&=\nabla\Big(\frac{\nabla D}{\sqrt{1+|\nabla D|^2}}\Big)\\&=\nabla\frac1{\sqrt{1+|\nabla D|^2}}\otimes\nabla D+\frac1{\sqrt{1+|\nabla D|^2}}\nabla^2D,
\end{split}
\label{shape-operator}
\end{equation*}
where 
$\nabla^2 D$ is the Hessian of $D$ and $\otimes$ is the Euclidean outer product. Regularizing the sparsity of the shape operator helps balance smoothness and geometric detail, reducing deformations caused by excessive bending or noise \cite{zhong2022spatially}. Thus, we apply a shape operator sparsity prior to the depth map yielding $P(D) \varpropto \exp\big(-\mathcal S(D)\big)$.
Additionally, we assume the albedo map $I$ follows a total variation (TV) prior to preserve edges and reduce noise. Consequently, we formulate the following energy functional to recover both the albedo and depth maps
\begin{equation*}
\begin{aligned}
E_2(u,D,I)= &\lambda \mathcal M\big(u,P^{\dagger}(I,D)\big)+ \eta\int_{\Upsilon} |\nabla I|dx\\&\quad+\int_{\Upsilon}\big|\alpha(D)\otimes\nabla D+\beta(D)\nabla^2D\big|_{F}dx,
\end{aligned}
\label{surface-recon}
\end{equation*}
where $\lambda$ and $\eta$ are positive parameters, $\mathcal M\big(u,P^{\dagger}(I,D)\big)$
is used to measure the distance between the hidden object and the back-projection surface in terms of the
$L_2$ norm, and $|\cdot|_F$ is the Frobenius norm. Besides, we denote
\[\alpha(D)=\nabla\frac{1}{\sqrt{1+|\nabla D|^2}},~~\beta(D)=\frac{1}{\sqrt{1+|\nabla D|^2}}.\]
\subsection{Geometric constrained NLOS reconstruction}
To address sparse illumination and low-exposure NLOS reconstruction, we jointly reconstruct the hidden object and measured surface, deriving the following model
\begin{equation}
  \min_{u,\tau,D,I} ~ E(u,\tau,D,I) := E_1(u,\tau) + E_2(u,D,I),
\label{joint-recon}
\end{equation}
which allows for higher-quality reconstruction by incorporating geometric surface constraints. 

\section{Solutions and Algorithm}

We now present the numerical minimization approach for the proposed model \eqref{joint-recon}, where the four variables are iteratively and alternatively solved as outlined in Algorithm \ref{alg1}. In our algorithm, the variable \( u \) is initialized by solving the following minimization problem
\begin{equation}
    \min_u ~~ \frac{1}{2} \int_\Omega (A_S u - \tau_0)^2 \, dx + \gamma \int_\Omega |u| \, dx,
\label{initialization of u}
\end{equation}
where \( A_S = SA \). This problem is solved using the Fast Iterative Shrinkage-Thresholding Algorithm (FISTA). Below, we will discuss the solution to each subproblem in detail.

\subsection{Sub-minimization problem w.r.t. the variable $\tau$}
The sub-minimization problem w.r.t. $\tau$ can be expressed as
\begin{equation*}
\min_\tau~~\frac12\int_\Omega(Au-\tau)^2dx+\frac{\rho}2\int_{\Omega}(S\tau-\tau_0)^2dx,
\label{tau problem}
\end{equation*}
which is a quadratic minimization problem. The close-form solution is given as follows
\begin{equation*}
\tau^{k+1}=\begin{cases}Au^{k},&~~ \mbox{for}~~S(x,y,t)=0,\\
(Au^{k}+\rho\tau_0)/(\rho+1),&~~ \mbox{for}~~S(x,y,t)=1,
\end{cases}
\label{tau0}
\end{equation*}
and $\tau_0$ is the sparse transient data.

\subsection{Sub-minimization problem w.r.t. the variable $D$}
According to $E_2$, the sub-minimization problem w.r.t the variable $D$ is defined as
\begin{equation*}
\min_D\frac12\int_{\Upsilon}(D-P_D(u))^2+\int_{\Upsilon}\big|\alpha(D)\otimes\nabla D+\beta(D)\nabla^2D\big|_{F}dx.
\label{problem D}
\end{equation*}
If we explicitly update $\alpha(D)$ and $\beta(D)$ as the weights for the first- and second-order gradient operators, the computational complexity can be effectively reduced, transforming the original problem into
\begin{equation}
\begin{aligned}
\min_D~\frac12 \int_{\Upsilon}(D-P_D(u))^2dx&+\int_\Upsilon|\alpha(D)||\nabla D|dx\\&+\int_\Upsilon\beta(D)|\nabla^2D|_Fdx.
\label{problem D3}
\end{aligned}
\end{equation}
By introducing auxiliary variable $v=\nabla D$ and $w=\nabla^2 D$, we can rewrite \eqref{problem D3} into a constrained minimization problem
\begin{equation*}
\begin{aligned}
&\min_{D,v,\omega}~~\frac{1}2\int_{\Upsilon}(D-P_D(u))^2+\int_{\Upsilon}\alpha(x)|v|+\int_{\Upsilon}\beta(x)|w|_{F}dx
\\&~~~\mathrm{s.t.}\quad v=\nabla D, w=\nabla^2D.
\end{aligned}
\label{problem Dvw}
\end{equation*}
Its augmented Lagrangian functional is defined as follows
\begin{equation}
\begin{aligned}
&\mathcal L(D,v,w;\Lambda_1,\Lambda_2)=\frac{1}2\int_{\Upsilon}(D-P_D(u))^2+\int_{\Upsilon}\alpha(x)|v|dx\\&-\int_{\Upsilon}\Lambda_1(v-\nabla D)+\frac{r_1}2\int_{\Upsilon}(v-\nabla D)^2+\int_{\Upsilon}\beta(x)|w|_{F}dx\\&-\int_{\Upsilon}\Lambda_2(w-\nabla^2 D)+\frac{r_2}2\int_{\Upsilon}(w-\nabla^2 D)^2dx,
\end{aligned}
\label{the augmented Lagrangian functionalD}
\end{equation}
where $\Lambda_1$ and $\Lambda_2$ are Lagrange multipliers, and $r_1$, $r_2$ are the positive penalty parameters. One immediately gets
the saddle-point problem below
\begin{equation*}\label{problem Dvw_L}    \max_{\Lambda_1,\Lambda_2}\min_{v, w, D}\mathcal L(v,w,D;\Lambda_1,\Lambda_2),
 \end{equation*}
which is further handled by alternative minimization w.r.t. the variables $v$, $w$ and $D$, respectively.

\subsubsection{Sub-minimization problem w.r.t. the variable $v$}
The sub-minimization problem w.r.t. the variable $v$ is defined as
 \begin{equation*}
\min_v\int_{\Upsilon}\alpha(x)|v|dx+\frac{r_1}{2}\int_{\Upsilon}\left(v-\nabla D-\frac{\Lambda_1}{r_1}\right)^2dx,
 \label{problem v}
 \end{equation*}
where $v^{k+1}$ can be computed through the  shrinkage operator as follows
\begin{equation*}
v^{k+1}=\mbox{shrinkage}\bigg(\nabla D^{k}+\frac{\Lambda_{1}^{k}}{r_1},\frac{\alpha(D^{k})}{r_1}\bigg)
\label{shrinkage v}
\end{equation*}
with $\mbox{shrinkage}(a,\xi)=\max\{|a|-\xi,0\}\frac a{|a|}$, for $a \in\mathbb{R}^n$.

\subsubsection{Sub-minimization problem w.r.t. the variable $w$}
The sub-minimization problem w.r.t. the variable $w$ is defined as follows
\begin{equation*}
\min_w \int_{\Upsilon} \beta |w|_F \, dx + \frac{r_2}{2} \int_{\Upsilon} \left(w - \nabla^2 D - \frac{\Lambda_2}{r_2}\right)^2  dx,
\label{problem_w}
\end{equation*}
where $w^{k+1}$ can also be computed by the shrinkage operator
\begin{equation*}
w^{k+1}=\mbox{shrinkage}_F\bigg(\nabla^2D^{k}+\frac{\Lambda_2^{k}}{r_2},\frac{\beta(D^{k})}{r_2}\bigg)
\label{shrinkage F}
\end{equation*}
with the shrinkage operator defined in terms of the Frobenius norm.


\subsubsection{Sub-minimization problem w.r.t. the variable $D$}
The sub-minimization problem w.r.t. the variable $D$ is defined as
\begin{equation*}
\begin{aligned}
\min_D~\frac{1}2\int_{\Upsilon}\Big(D-P_D(u)&\Big)^2+\frac{r_1}2\int_{\Upsilon}\Big(\nabla D-(v-\frac{\Lambda_1}{r_1})\Big)^2dx\\&+\frac{r_2}2\int_{\Upsilon}\Big(\nabla^2D-(w-\frac{\Lambda_2}{r_2})\Big)^2dx.
\end{aligned}
\label{problem D4}
\end{equation*}
Given the variables $v^{k+1},\omega^{k+1}, \Lambda_1^k, \Lambda_2^k$,
the Euler-Lagrange equation of the above subproblem can be obtained by
\begin{equation*}
\begin{aligned}
D-P_D(u^k)-&r_1\text{div}\Big(\nabla D-(v^{k+1}-\frac{\Lambda_1^k}{r_1})\Big)
  \\&+r_2\text{div}^2\Big(\nabla^2D-(w^{k+1}-\frac{\Lambda_2^k}{r_2})\Big)=0.
\end{aligned}
\end{equation*}
We can solve the above PDE by Fast Fourier Transform (FFT), which gives
\begin{equation*}
\begin{aligned}
&D^{k+1}=\\
&\mathcal{F}^{-1}\bigg(\frac{\mathcal{F}\big(P_D(u^k)-\text{div}(r_1v^{k+1}-\Lambda_1^{k})+\text{div}^2(r_2w^{k+1}-\Lambda_2^{k})\big)}{\big(\mathcal{I}-r_1\mathcal{F}\Delta\mathcal{F}^{-1}+r_2\mathcal{F}\Delta^2\mathcal{F}^{-1}\big)}\bigg),
\end{aligned}
\label{FFT D}
\end{equation*}
where $\mathcal F$ and $\mathcal F^{-1}$ represent the forward and inverse FFT, respectively.

\subsubsection{Update of $\alpha$ and $\beta$} Both \(\alpha\) and \(\beta\) are adaptively updated according to the following rules
\begin{equation*}
\begin{cases}\alpha(D^{k+1})=\nabla\frac{1}{\sqrt{1+|\nabla D^{k+1}|^2}},\\
\beta(D^{k+1})=\frac{1}{\sqrt{1+|\nabla D^{k+1}|^2}}.
\end{cases}
\label{alpha and beta}
\end{equation*}

\subsubsection{Update of Lagrange multipliers}
The initial values of the Lagrange multipliers $\Lambda_1$ and $\Lambda_2$ are set to zero. They are updated through a standard dual-ascent rule from:
\begin{equation*}
\begin{cases}\Lambda_1^{k+1}=\Lambda_1^k+r_1(\nabla D^{k+1}-v^{k+1}),\\
\Lambda_2^{k+1}=\Lambda_2^k+r_2(\nabla^2D^{k+1}-\omega^{k+1}).
\end{cases}
\label{lambda1and2}
\end{equation*}


\subsection{Sub-minimization problem w.r.t. $I$}
The minimization problem w.r.t. the albedo function $I$ is given as follows
\begin{equation*}
\min_I~\frac12\int_{\Upsilon}(I-P_I(u))^2dx+\int_{\Upsilon}\eta|\nabla I|dx,
\label{problem $I$}
\end{equation*}
which is a typical total variation minimization problem.  By introducing an auxiliary variable $q=\nabla I$, we can rewrite
the above equation into a constrained minimization problem as
\begin{equation*}
\begin{aligned}
\min_{I,q}\quad&\frac{1}2\int_{\Upsilon}(I-P_I(u))^2dx+\int_{\Upsilon}\eta|q|dx
\\\mathrm{s.t.}\quad &q=\nabla I,
\end{aligned}
\label{problem I1}
\end{equation*}
and the corresponding augmented Lagrangian is as follows
\begin{equation*}
\begin{aligned}
\mathcal L(I,q;\Lambda_3)=&\frac{1}2\int_{\Upsilon}(I-P_D(u))^2dx +\int_{\Upsilon}\eta|q|dx\\
&-\int_{\Upsilon}\Lambda_3(q-\nabla I)dx+\frac{r_3}{2}\int_{\Upsilon}(q-\nabla I)^2dx.
\end{aligned}
\label{the augmented Lagrangian functionalI}
\end{equation*}
Similarly, the saddle point problem is formulated as
\begin{equation*}\label{problem Iq}
    \max_{\Lambda_3}\min_{q, I}\mathcal L(q,I;\Lambda_3),
 \end{equation*}
which can be further split into two sub-minimization problems w.r.t. $q$ and $I$, respectively.

\begin{figure*}[t]
       \begin{center}			
       \includegraphics[width=0.95\linewidth]{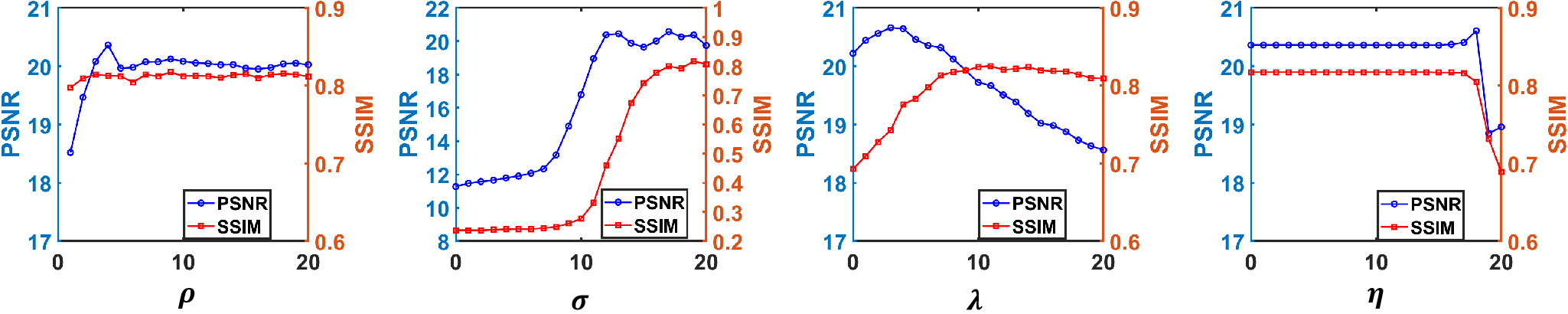}
 	  \end{center}
      \vspace{-0.2cm}
 	\caption{
Performance of our SLCT model in terms of both SSIM and PSNR with different model parameters $\rho$, $\sigma$, $\lambda$, and $\eta$. }
 	\label{para2}
 \end{figure*}

\subsubsection{Sub-minimization problem w.r.t. the variable $q$}
 The sub-minimization problem w.r.t. $q$ is defined as
 \begin{equation*} \min_q\int_{\Upsilon}\eta|q|dx+\frac{r_3}2\int_{\Upsilon}\Big(q-\nabla I-\frac{\Lambda_3}{r_3}\Big)^2dx,
 \label{problem q}
 \end{equation*}
which can be solved through the soft thresholding operator
\begin{equation*}
q^{k+1}=\mbox{shrinkage}\Big(\nabla I^k+\frac{\Lambda_3^k}{r_3},\frac{\eta}{r_3}\Big).
\label{shrinkage q}
\end{equation*}

\subsubsection{Sub-minimization problem w.r.t. the variable $I$}
The sub-minimization problem w.r.t. $I$ is defined as
 \begin{equation*}
 \min_I~\frac{1}2\int_{\Upsilon}(I-P_I(u))^2dx+\frac{r_3}2\int_{\Upsilon}\Big(\nabla I-(q-\frac{\Lambda_3}{r_3})\Big)^2dx.
 \label{problem I2}
 \end{equation*}
Given the fixed variables $q^{k+1}$ and $\Lambda_3^k$, the relevant Euler-Lagrange equation is defined as
 \begin{equation*}
 I-P_I(u^k)-r_3{\text{div}}\Big(\nabla I-(q^{k+1}-\frac{\Lambda_3^k}{r_3})\Big)=0,
 \label{problem I3}
 \end{equation*}
which can be solved by FFT as follows
\begin{equation*}
I^{k+1}=\mathcal{F}^{-1}\bigg(\frac{\mathcal{F}\big(P_I(u^k)-\text{div}(r_3q^{k+1}-\Lambda_3^k)\big)}{\mathcal{I}-r_3\mathcal{F}\Delta\mathcal{F}^{-1}}\bigg).
\label{FFT I}
\end{equation*}

\subsubsection{Update of Lagrange multipliers}

The initial value $\Lambda_3^0=0$ and it can be updated from
\begin{equation*}
\Lambda_3^{k+1}=\Lambda_3^k+r_3(\nabla I^{k+1}-q^{k+1}).
\label{lambda3}
\end{equation*}

\subsection{Sub-minimization problem w.r.t. $u$}
Since both energy functional $E_1$ and $E_2$ contain the variable $u$, the sub-minimization problem w.r.t. $u$ can be given as
\begin{equation*}
\min_u~\frac12\int_\Omega(Au-\tau)^2+\frac{\lambda}2 \int_{\Omega}(u-P^{\dagger}(I,D))^2+\gamma\int_\Omega|u|dx,
\label{problem u}
\end{equation*}
which can be solved by the Fast Iterative Shrinkage-Thresholding Algorithm (FISTA) \cite{beck2009fast}. Thus,  $u^{k+1}$ is updated by
\begin{equation*}
\begin{aligned}
u^{k+1}=\mbox{shrinkage}\Big(\overline{u}^k&-t[A^T(A\overline{u}^k-\tau^{k+1})\\&+\lambda(\overline{u}^k-P^{\dagger}(I^{k+1},D^{k+1})],\gamma t\Big),
\end{aligned}\label{shrinkage u}
\end{equation*}
where $\overline{u}$ is an extrapolation variable used to accelerate updates. Initially, $\overline{u}^{0}=u^0$ and $\mu^0=1$. The extrapolation variable is updated iteratively using
 \begin{equation*}
\begin{cases}\overline{u}^{k+1}=u^{k+1}+\frac{\mu^k-1}{\mu^{k+1}}(u^{k+1}-u^k),\\
\mu^{k+1}=(1+\sqrt{1+4(\mu^k)^2})/2.
\end{cases}
\label{u_bar}
\end{equation*}
The iterative step size $t$ is chosen to balance the algorithm's convergence and convergence speed. A suitable choice for $t$ is $1/(||A||_2^2+\lambda)$.



\section{Experimental results}
\label{sec:results}

All our experiments are conducted on a workstation equipped with an Intel Xeon Gold 6132 CPU and an NVIDIA TITAN RTX GPU. Both the CNLOS method and our approach are GPU-accelerated and implemented on the same GPU. To illustrate the effectiveness of our proposed method in improving both the quality and efficiency of NLOS imaging reconstruction. We compare it with existing methods (i.e., LCT \cite{o2018confocal}, DLCT \cite{young2020non}, PF \cite{liu2020phasor},  SPIRAL \cite{ye2021compressed}, SSCR \cite{liu2023few}, CNLOS \cite{ding2024curvature}). Experimental validation is implemented on the Zaragoza\cite{galindo2019dataset} ,
 Stanford\cite{lindell2019wave} , USTC\cite{li2023nlost}  and NLoS benchmark\cite{klein2018quantitative} datasets.

\subsection{Data description and preprocessing}
The datasets `usaf', `serapis', and `bunny' are from the Zaragoza dataset, with original dimensions of 256×256, 256×256, and 64×64, and wall sizes of 1m×1m, 1m×1m, and 0.6m×0.6m, respectively. Each scenario has 512 time bins, with photons traveling 0.003m, 0.003m, and 0.0025m per bin. We preprocess the data by uniformly sampling and resizing it to 64×64×512. The datasets `statue', `dragon', `outdoor', and `teaser' are from the Stanford dataset. The raw measurement size for `outdoor' is 128×128×2048, while the others are 512×512×2048, all with a wall size of 2m×2m. The time resolution is cropped to 512 bins, each spanning 32 ps. `Statue', `dragon', and `outdoor' are uniformly sampled and resized to 64×64. According to \cite{li2023nlost}, `teaser' is preprocessed to 256×256 and further reduced to 128×128 by aggregating every 2×2 points. The `man letter' dataset is from the USTC dataset, with dimensions of 128×128×512, a wall size of 2m×2m, and each time bin spanning 32 ps. The non-confocal data `k' is from the NLOS benchmark dataset, with a wall size of 0.256m×0.256m and photons traveling 0.001m per bin. For our experiments, we use data downsampled to 64×64, as provided by \cite{liu2021non}. It includes 1024 time bins.

\subsection{Parameter discussing}

Our model includes six parameters: \(\rho\), \(\sigma\), \(\lambda\), \(\eta\), \(\alpha\), and \(\beta\), where \(\alpha\) and \(\beta\) are adaptively updated based on depth values. The algorithm also incorporates penalty parameters \(r_1\), \(r_2\), and \(r_3\), with appropriate values accelerating convergence. Since most experiments are performed on undersampled measurements, we evaluate the impact of these parameters on the dragon scene at a 16×16 scanning points.


   \begin{figure*}[t]
       \begin{center}			
       \includegraphics[width=0.76\linewidth]{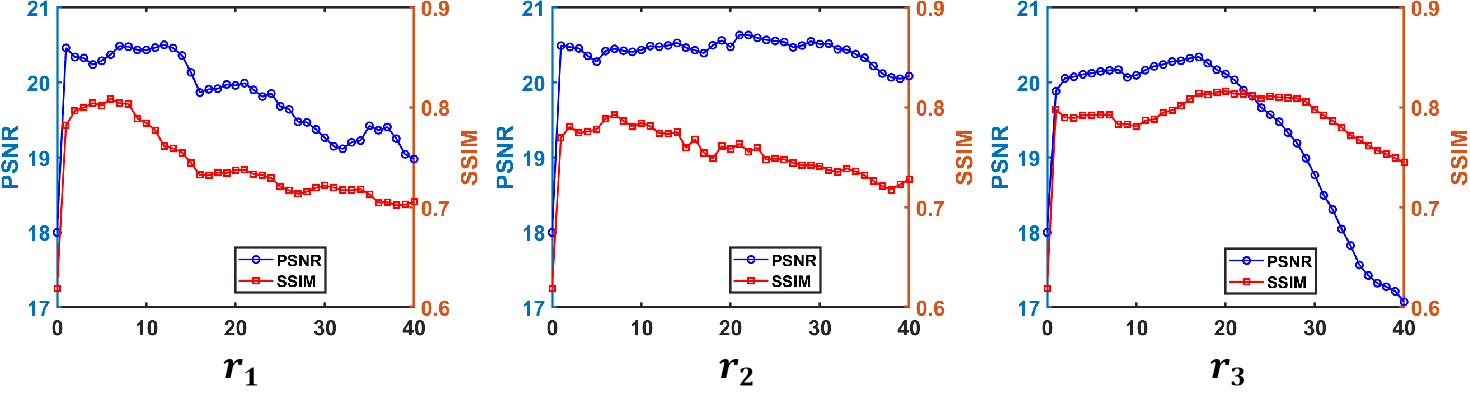}
 	  \end{center}
      \vspace{-0.2cm}
 	\caption{
Performance of our SLCT model in terms of both SSIM and PSNR with different algorithm parameters $r_1$, $r_2$, and $r_3$. }
 	\label{para1}
 \end{figure*}

\begin{figure}[t]
       \begin{center}			
       \includegraphics[width=0.95\linewidth]{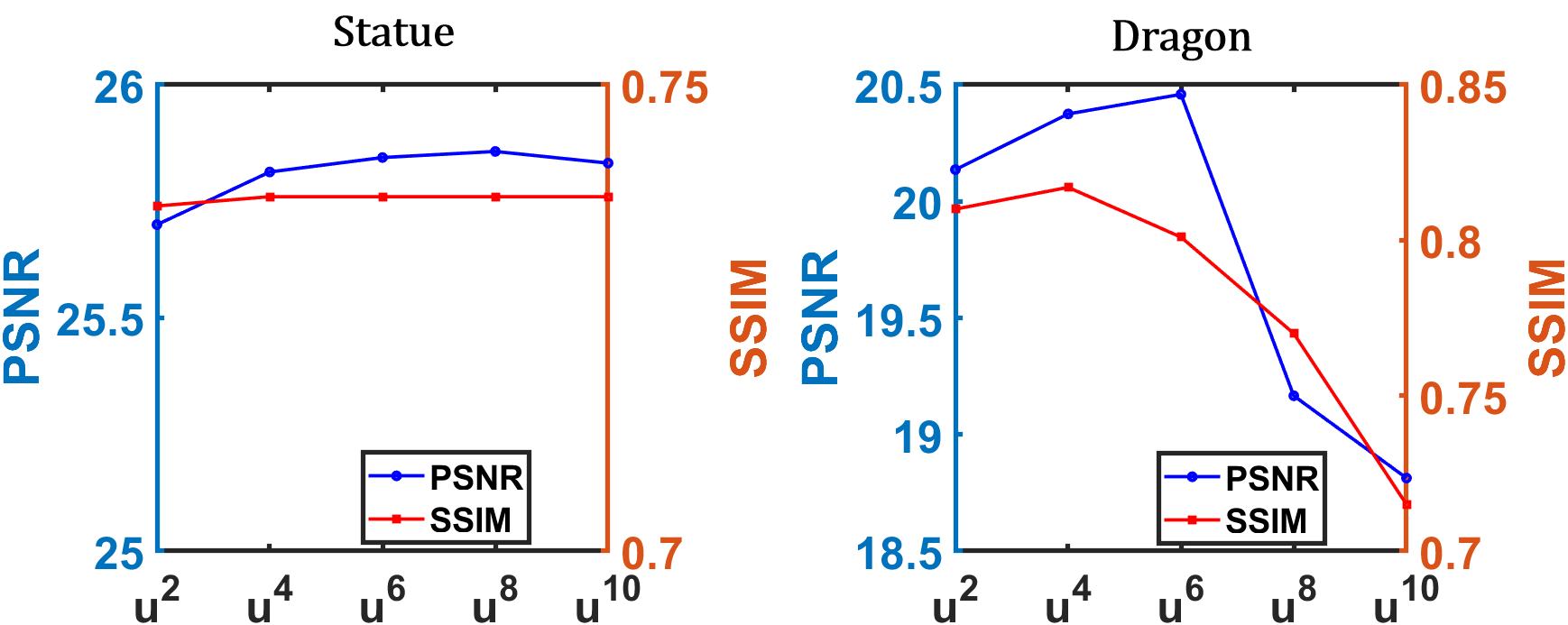}
 	  \end{center}
 	\caption{Evaluation of model performance with different powers of $u$.}
 	\label{para3}
     \vspace{-0.5cm}
 \end{figure}

Fig. \ref{para2} illustrates the impact of varying model parameters on the results. The ranges for \(\rho\), \(\sigma\), and \(\lambda\) are [0, 100], [0, 1e-5], and [0, 1], respectively, with 20 uniformly selected values tested for each. For \(\eta\), ranging from [0, 1e-1], 20 logarithmically spaced values are tested. The results indicates that our model is less sensitive to \(\rho\) and \(\eta\) within certain ranges, which can be fixed as \(\rho = 25\) and \(\eta = 1e-5\). However, our model is sensitive to \(\sigma\) and \(\lambda\), necessitating careful case-by-case tuning. The algorithm parameters \(r_1\), \(r_2\), and \(r_3\) have ranges of [0, 0.8], [0, 8], and [0, 40], respectively, with 40 uniformly selected values tested. As shown in Fig. \ref{para1}, all algorithm parameters can also be fixed, specifically \(r_1 = 0.1\), \(r_2 = 2\), and \(r_3 = 20\).
Additionally, we investigate the impact of the power $p$ in  the weight calculation \eqref{weight}. While a higher power emphasizes larger \(u\) values, it does not always improve results. As shown in Fig. \ref{para3}, in the dragon scenario, high weights may amplify noise, resulting in a significant decrease in the SSIM value. Thus, we fix the power $p=4$.

\begin{figure}[t]
       \begin{center}			
       \includegraphics[width=0.95\linewidth]{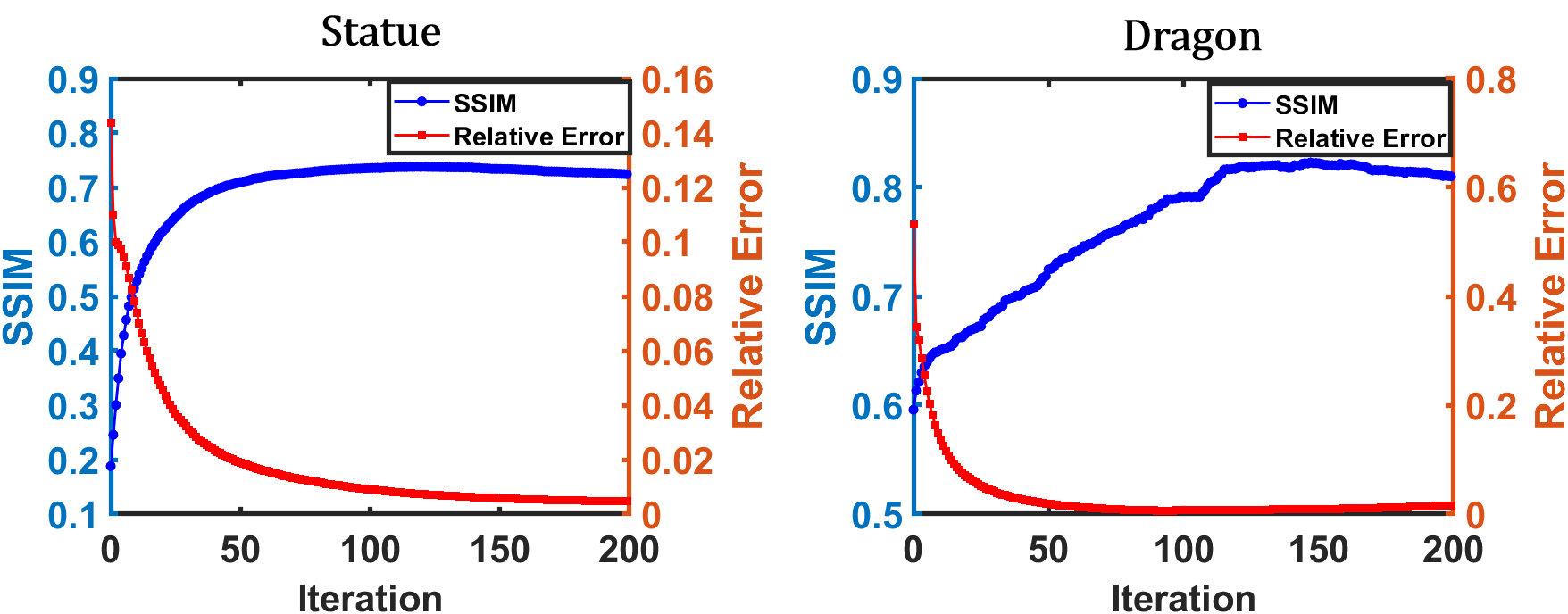}
 	  \end{center}
 	\caption{Relative error $|u^{k+1} - u^k|/|u^k|$ and SSIM w.r.t. different numbers of iterations.}
 	\label{para4}
    \vspace{-0.5cm}
 \end{figure}

We also discuss the algorithm's iteration count in two different scenarios. As shown in Fig. \ref{para4}, with an increase in iterations, the relative error decreases and the metric values stabilize. We consider both the time cost and the quality of the image reconstruction and set the maximum number of iterations to 120.

\subsection{Comparison among LCT, DLCT and our SLCT}
\begin{figure*}[h]
      \begin{center}			
      \includegraphics[width=0.85\linewidth]{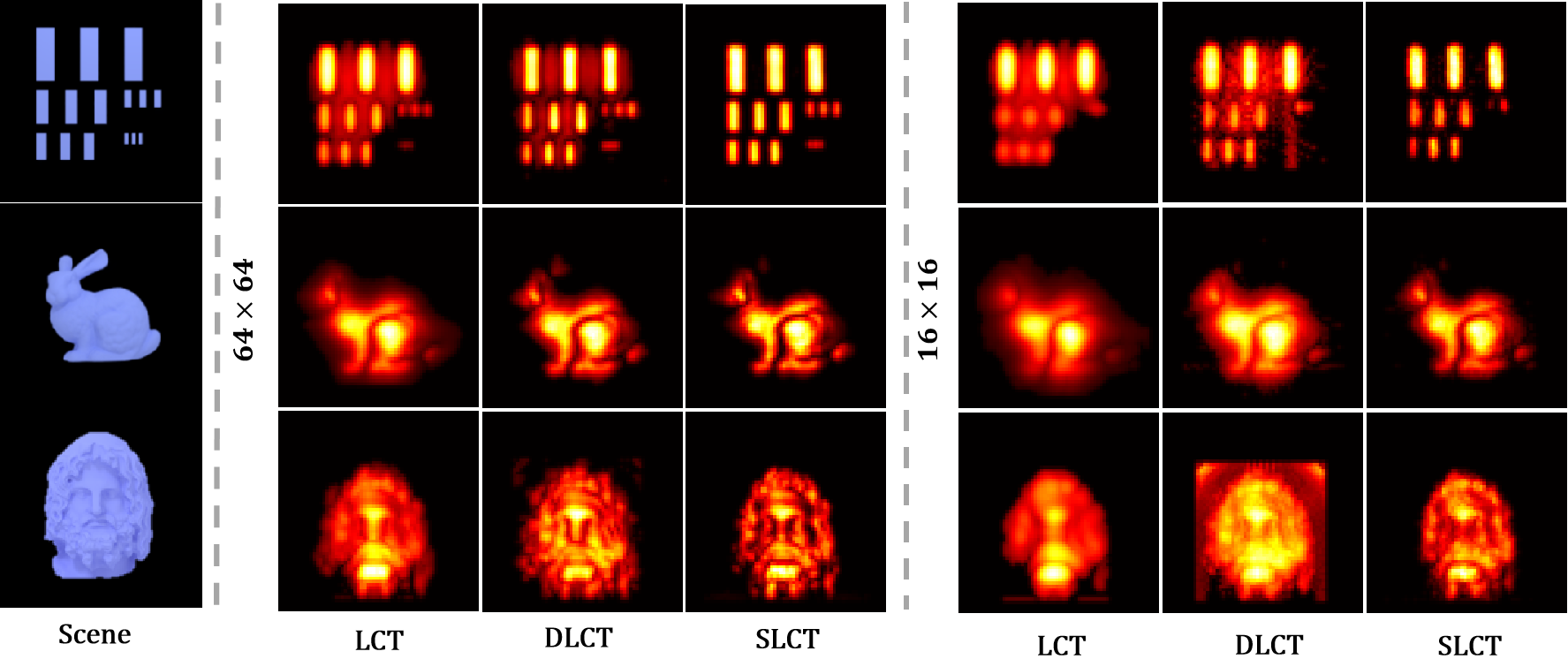}
	  \end{center}
	\caption{Comparison results on `usaf', `bunny' and `serapis' with 64$\times$64 and 16$\times$16  scanning points. The parameters of our methods are set as: $\sigma = 5e-7$, $\lambda = 0.02$ (usaf), $\sigma = 5e-7$, $\lambda = 0.02$ (bunny),  $\sigma = 5e-7$, $\lambda = 0.02$ (serapis) for 64$\times$64 scanning points; $\sigma = 8e-6$, $\lambda = 0.02$ (usaf), $\sigma = 1e-6$, $\lambda = 0.04$ (bunny),  $\sigma = 4e-6$, $\lambda = 0.03$ (serapis) for 16$\times$16 scanning points.  }
	\label{fig8}
\end{figure*}

We conduct a qualitative comparison of LCT and DLCT on the Zaragoza dataset. As shown in Fig. \ref{fig8}, our SLCT method achieves high-quality reconstruction across all scenes. With sufficient illumination points, SLCT provides more accurate reconstruction results in the first two scenes and performs comparably to DLCT in the serapis scene. However, as the number of illumination points decreases, the performance of LCT and DLCT significantly degrades, particularly in the detail-rich serapis scene, where DLCT generates severe background noise that cannot be effectively removed through post-processing. This indicates that the normal vector estimation errors in DLCT increase significantly with fewer illumination points and higher noise levels, leading to a decline in reconstruction quality. In contrast, our method consistently reconstructs superior shapes and details under sparse illumination conditions, demonstrating stronger robustness.

\begin{figure}[t]
      \begin{center}			
      \includegraphics[width=0.95\linewidth]{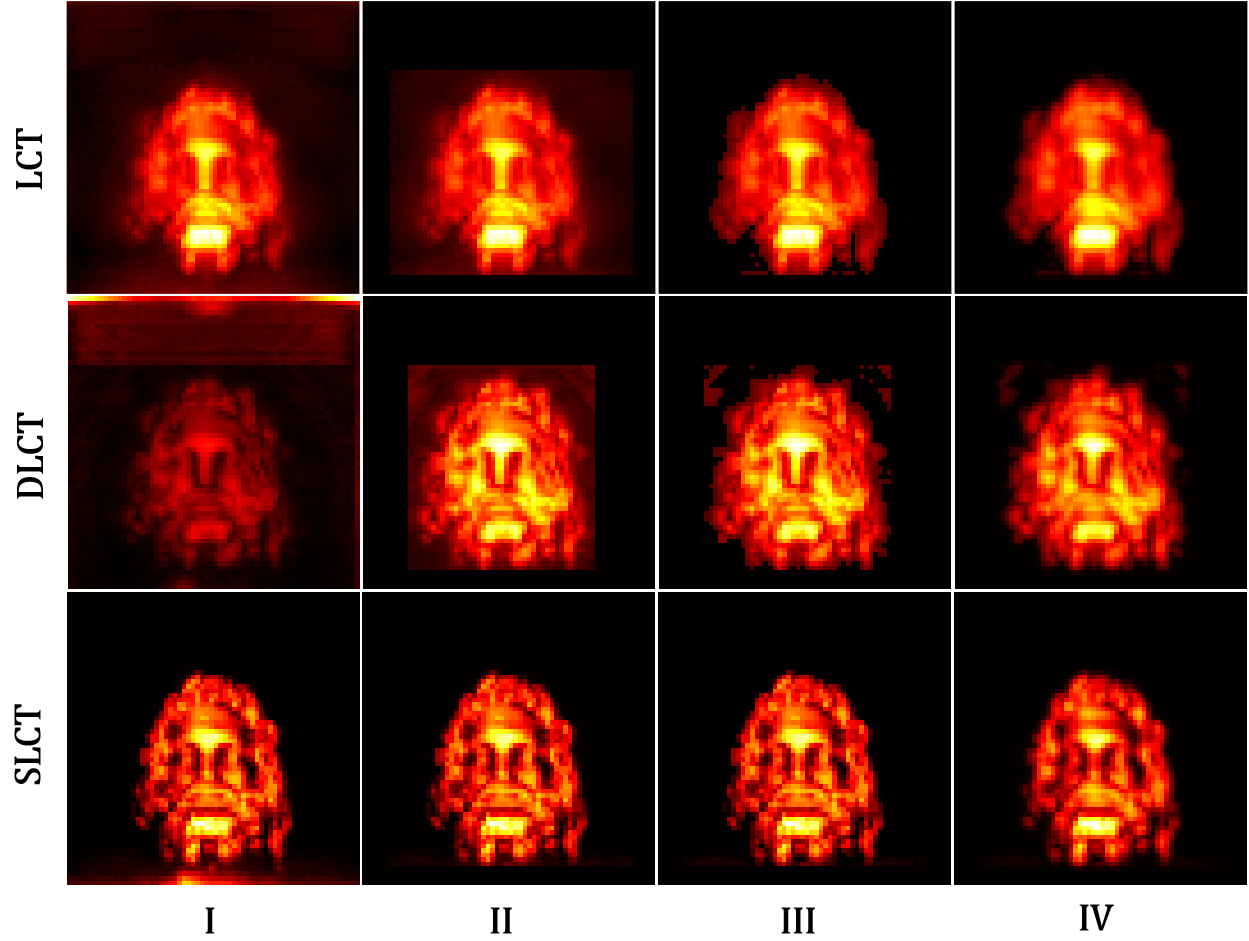}
	  \end{center}
	\caption{The impact of post-processing methods on the results. I: No post-processing; II: Truncation post-processing; III: Truncation and thresholding post-processing; IV: Truncation, thresholding and Gaussian smoothing post-processing.}
	\label{fig7}
\vspace{-0.5cm}
\end{figure}

Additionally, we investigate the impact of post-processing on non-line-of-sight imaging results. Three common post-processing techniques are employed across different methods to enhance reconstruction quality: truncation removes outliers, thresholding eliminates smaller values, and Gaussian smoothing refines the image. Fig. \ref{fig7} demonstrates the effects of these techniques on the reconstruction of the serapis scene under full sampling conditions. Evidently, compared to our method, direct approaches such as LCT and DLCT exhibit a stronger dependence on post-processing. In subsequent experiments, we apply appropriate post-processing to the results of different methods. As individually adjusting post-processing for each image is time-consuming and inefficient for non-line-of-sight imaging, in later real-data experiments, we use a consistent post-processing approach for the same method across different scanning points within the same scene.

\subsection{Comparison results on real measurement data}

\begin{figure*}[ht]
    \begin{center}			
      \includegraphics[width=0.89\linewidth]{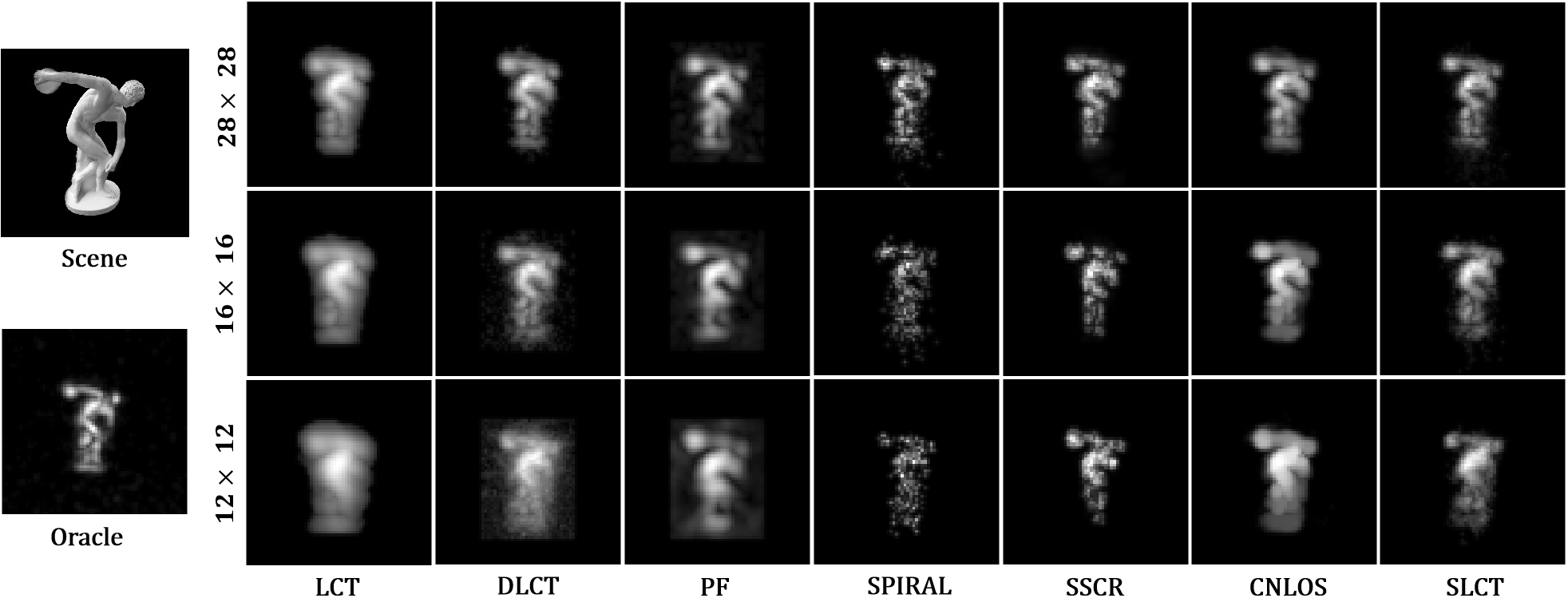}
	  \end{center}
	\caption{Comparison results on `statue' with exposure time of 60 min, 
    where the oracle is obtained by PF using 64 $\times$ 64 scanning points with exposure time of 180 min. The parameters of our methods are set as: $\sigma = 7.5e-6$, $\lambda = 0.02$ (28 $\times$ 28); $\sigma = 8e-6$, $\lambda = 0.05$ (16 $\times$ 16);  $\sigma = 1e-5$, $\lambda = 0.08$ (12 $\times$ 12). }
	\label{fig9}
\end{figure*}

We present reconstruction results for the `statue' scene using various methods. As shown in Fig. \ref{fig9}, direct methods such as DLCT and PF produce high-quality results with a higher number of scanning points. However, as the number of illumination points decreases, these methods show increased background noise compared to iterative approaches, with all methods experiencing performance degradation. In contrast, our SLCT method captures more details even with only 12×12 scanning points. Table \ref{table:statue time} provides evaluation metrics, demonstrating that SLCT consistently achieves higher PSNR and SSIM values across all scenes. Our SLCT achieves higher reconstruction accuracy with only 16×16 scanning points than direct methods at 28×28 scanning points, while requiring less total time than the fastest direct method. This highlights the importance of reducing scanning points to accelerate NLOS reconstruction.

\begin{table*}[t]
\footnotesize
\setlength{\tabcolsep}{4pt}
\renewcommand{\arraystretch}{1.2}
\caption{Comparison w.r.t. different numbers of illumination points on the `statue' with an exposure time of 60 min, where `R time' and `T time' represent reconstruction and total time (data acquisition and reconstruction) in seconds, respectively. }
\label{table:statue time}
\centering
\begin{tabular}{c| c c c c|c c c c|c c c c}
\toprule
\multirow{2}{*}{\textbf{Methods}} & \multicolumn{4}{c|}{\textbf{28$\times$28}} & \multicolumn{4}{c|}{\textbf{16$\times$16}} & \multicolumn{4}{c}{\textbf{12$\times$12}} \\
\cline{2-13}
 & \textbf{PSNR$\uparrow$} & \textbf{SSIM$\uparrow$} & \textbf{Rtime$\downarrow$} & \textbf{Ttime$\downarrow$}
 & \textbf{PSNR$\uparrow$} & \textbf{SSIM$\uparrow$} & \textbf{Rtime$\downarrow$} & \textbf{Ttime$\downarrow$}
 & \textbf{PSNR$\uparrow$} & \textbf{SSIM$\uparrow$} & \textbf{Rtime$\downarrow$} & \textbf{Ttime$\downarrow$} \\
\midrule
LCT\cite{o2018confocal} & 20.915  & 0.668  & 0.5 & 11.3 & 19.668 & 0.633 & 0.5 & 4.0 & 17.780 & 0.588 & 0.5 & 2.5 \\
DLCT\cite{young2020non} & 22.848  & 0.682  & 5.8 & 16.6 & 22.228 & 0.625 & 5.8 & 9.3 & 18.588 & 0.534 & 5.8 & 7.8 \\
PF\cite{liu2020phasor} & 23.939  & 0.632  & 0.9 & 11.7& 23.008 & 0.602 & 0.9 & 4.4 & 20.357 & 0.547 & 0.9 & 2.9 \\
SPIRAL\cite{ye2021compressed}   &24.242   &0.703   &34.2  &45.0  &21.918  &0.676  &43.3  &46.8  &20.690  &0.646  &39.5  &41.5  \\
SSCR\cite{liu2023few}  &25.527  & 0.733  &1469.5  &1480.3 &23.616  &0.695  &490.6  &494.1  &22.654  &0.675  &268.7  &270.7  \\
CNLOS\cite{ding2024curvature} &25.318   &0.711   &15.2  &26.0  &23.006  &0.688  &15.2  &18.7  &20.849  &0.669  &15.2  &17.2 \\
\midrule
\textbf{SLCT}  & \textbf{25.813} &\textbf{0.738} & \textbf{5.4} &\textbf{16.2} & \textbf{24.304} & \textbf{0.710} & \textbf{5.4} & \textbf{8.9} & \textbf{22.969} & \textbf{0.682} & \textbf{5.4} & \textbf{7.4} \\
\bottomrule
\end{tabular}
\end{table*}

\begin{figure*}[t]
      \begin{center}			
      \includegraphics[width=0.89\linewidth]{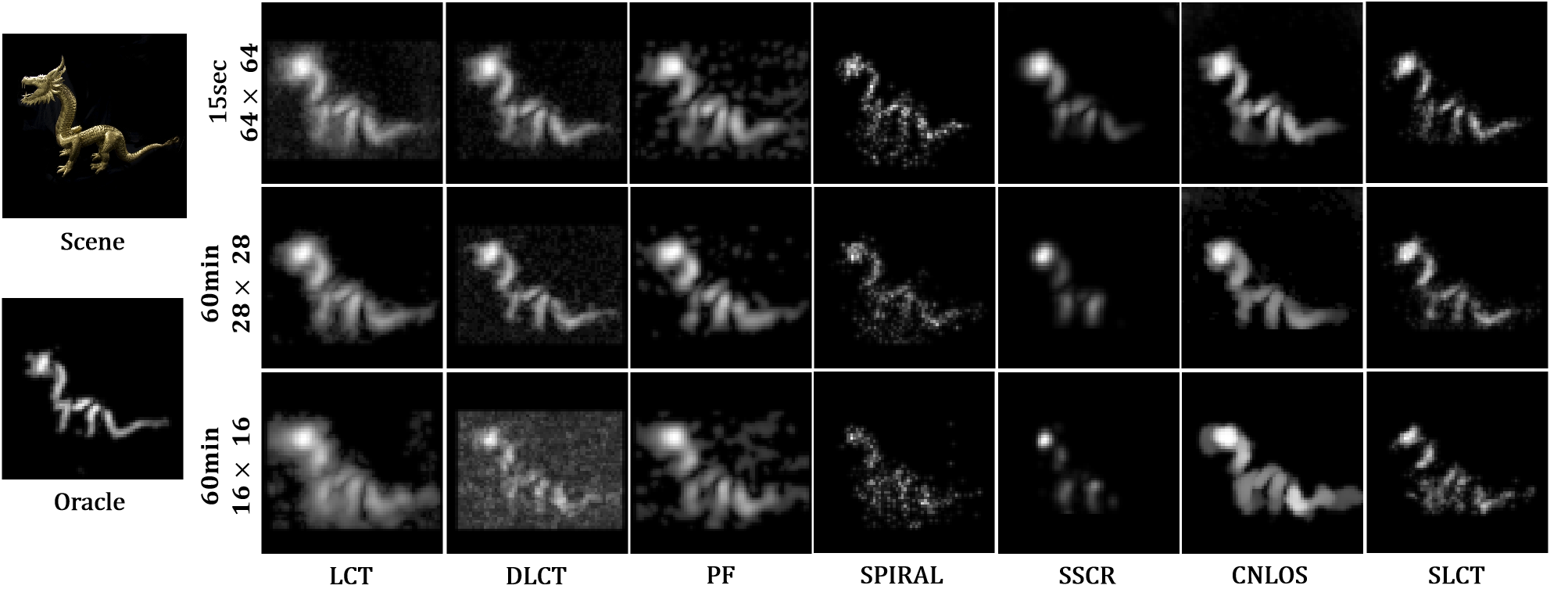}
	  \end{center}
	\caption{Comparison of reconstruction results using different methods on `dragon'  with varying scanning points and exposure times. The oracle is obtained by PF using 64 $\times$ 64 measurements of exposure time 180 min. The parameters of our methods are set as: $\sigma = 8e-6$, $\lambda = 0.1$ (15sec, 64 $\times$ 64); $\sigma = 8e-6$, $\lambda = 0.1$ (60min, 28 $\times$ 28);  $\sigma = 9.5e-6$, $\lambda = 0.35$ (60min, 16 $\times$ 16).}
	\label{iterative-compare-dragon}
\end{figure*}

\begin{figure*}[ht]
       \begin{center}			
       \includegraphics[width=0.89\linewidth]{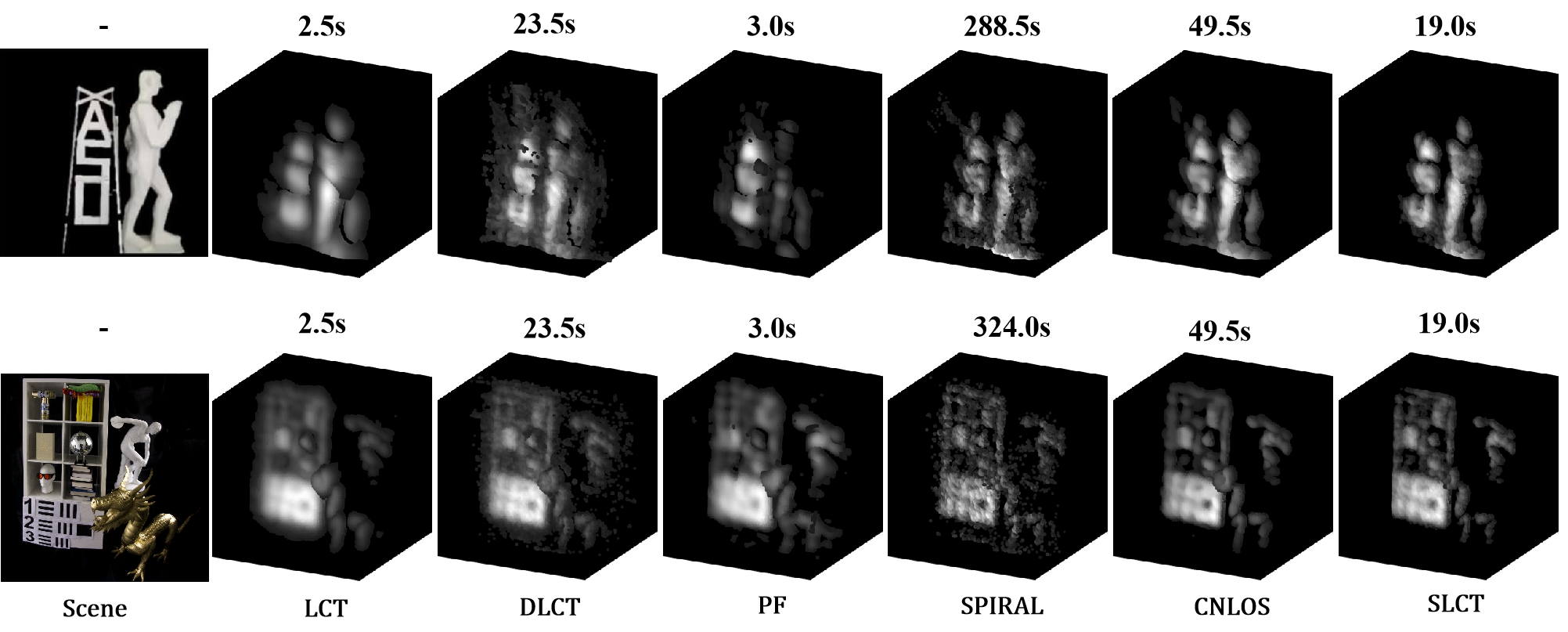}
 	  \end{center}
 	\caption{Comparison results on `man letter' and `teaser' with 16$\times$16 scanning points and 28$\times$28 scanning points, respectively. The parameters of our methods are set as: $\sigma = 8e-6$, $\lambda = 0.12$ (man letter); $\sigma = 6.5e-6$, $\lambda = 0.1$ (teaser).}
 	\label{complex-scene}
 \end{figure*}

\begin{table*}[t]
\footnotesize
\setlength{\tabcolsep}{4pt}
\renewcommand{\arraystretch}{1.2}
\caption{Comparison among different methods with various scanning points and exposure times on `dragon', where time is recorded in seconds unless otherwise specified.}
\label{table:dragon metrics}
\centering
\begin{tabular}{c| c c c c|c c c c|c c c c}
\toprule
\multirow{2}{*}{\textbf{Methods}} & \multicolumn{4}{c|}{\textbf{15sec~64$\times$64}} & \multicolumn{4}{c|}{\textbf{60min~28$\times$28}} & \multicolumn{4}{c}{\textbf{60min~16$\times$16}} \\
\cline{2-13}
 & \textbf{PSNR$\uparrow$} & \textbf{SSIM$\uparrow$} & \textbf{Rtime$\downarrow$} & \textbf{Ttime$\downarrow$}
 & \textbf{PSNR$\uparrow$} & \textbf{SSIM$\uparrow$} & \textbf{Rtime$\downarrow$} & \textbf{Ttime$\downarrow$}
 & \textbf{PSNR$\uparrow$} & \textbf{SSIM$\uparrow$} & \textbf{Rtime$\downarrow$} & \textbf{Ttime$\downarrow$} \\
\midrule
LCT\cite{o2018confocal} &18.238   &0.416  &0.5  &0.7  &19.462  &0.711  &0.5  &11.3  &16.041  &0.477  &0.5  &4.0  \\
DLCT\cite{young2020non} &20.367   &0.488   &5.8  &6.0  &19.725  &0.424  &5.8  &16.6  &15.055 &0.331  &5.8  &9.3  \\
PF\cite{liu2020phasor} &19.225   &0.453   &0.9  &1.1  & 20.110 &0.641  &0.9  &11.7  &18.108  &0.474  &0.9  &4.4  \\
SPIRAL\cite{ye2021compressed}   & 21.408  & 0.791  & 30.5 & 30.7 & 20.506 & 0.771 & 27.7 & 38.5 & 19.639 & 0.742 & 24.9 & 28.4 \\
SSCR\cite{liu2023few}  & 21.749       & 0.838      & $>40 min$ & $>40 min$ & 18.760 & 0.781 & 978.9 & 989.7 & 18.236 & 0.774 & 408.1 &411.6  \\
CNLOS\cite{ding2024curvature} & 22.680  & 0.733  & 15.2 & 15.4 & \textbf{22.227} & 0.728 & 15.2 & 26.0 & 18.325 & 0.771 & 15.2 & 18.7 \\
\midrule
\textbf{SLCT}  & \textbf{23.265} &\textbf{0.870} & \textbf{5.4} &\textbf{5.6} & 22.119& \textbf{0.841} & \textbf{5.4} & \textbf{16.2} & \textbf{20.375} & \textbf{0.817} & \textbf{5.4} & \textbf{8.9} \\
\bottomrule
\end{tabular}
\end{table*}

 We compared various methods on the `dragon' scene under different exposure times and scanning points. The dragon's metallic surface introduces more light scattering noise than the `statue' scene, increasing background noise for direct methods. SPIRAL, an iterative method, produces rough reconstructions but is highly noise-sensitive. SSCR also suffers from noise, leading to significant albedo loss. While CNLOS excels on smooth surfaces, its strong continuity priors for 3D curvature and volumetric reflectance result in fine detail loss. In contrast, SLCT achieves the highest reconstruction resolution. Table \ref{table:dragon metrics} provides quantitative metrics, validating the robustness of our approach. Our method achieves comparable accuracy while significantly reducing processing time, offering over 30× speedup compared to SSCR and reducing reconstruction time to one-third of GPU-accelerated CNLOS.

We compared our method with other established approaches in complex scenes. As shown in Fig. \ref{complex-scene}, LCT and PF are limited to reconstructing only the outlines of hidden objects. DLCT captures subtle details by incorporating normal vectors, but this results in significant noise. Iterative methods, however, reveal the fine structure of objects and identify specific features. Specifically, CSA introduces noticeable background noise and albedo loss, while CNLOS produces overly smooth reconstructions, particularly at the boundaries. Our SLCT method effectively reduces noise and better preserves texture, achieving the highest resolution. Additionally, our method offers the fastest reconstruction time among iterative methods.

\subsection{Ablation studies}

In Table \ref{table:ablation}, we conduct ablation studies using real measurement data from the Stanford dataset to validate the effectiveness of our model. Setup I serves as the baseline, incorporating only \(L_1\) regularization on the reconstructed image. Setup II enhances this by adding a transient data inpainting module, improving PSNR by 0.5 dB and 0.4 dB in two scenarios, albeit at the cost of reduced SSIM. Setup III, our proposed SLCT model, introduces geometric constraints, significantly improving both reconstruction accuracy and resolution, with notable gains in PSNR and SSIM. Fig. \ref{ablation-expe} demonstrates that SLCT achieves the most accurate shape reconstructions, particularly in the head and body of the person in the outdoor scene, as well as the dragon's body. Both qualitative and quantitative comparisons highlight that geometric constraints on the object surface substantially enhance NLOS reconstruction quality.

\subsection{Comparison results on nonconfocal data}
\begin{figure*}[ht]
       \begin{center}			
       \includegraphics[width=0.88\linewidth]{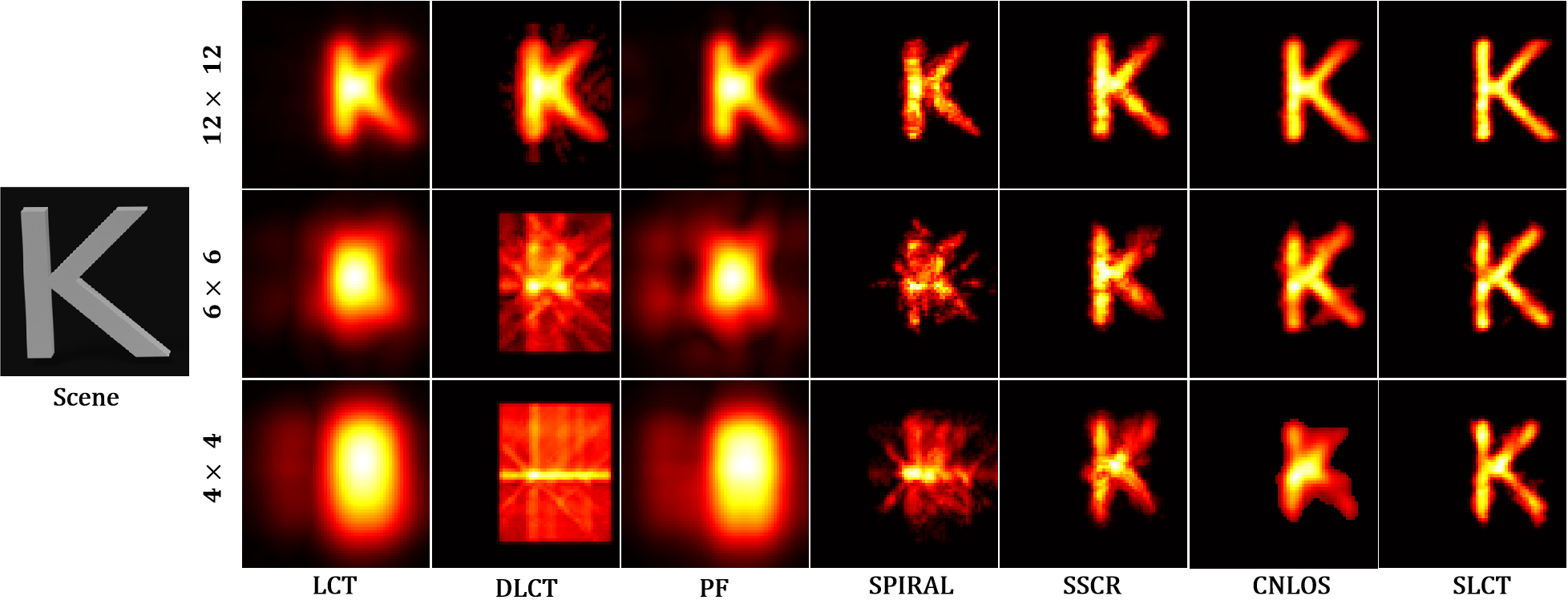}
 	  \end{center}
 	\caption{Comparison on the nonconfocal dataset `K' with 4×4, 6×6, and 12×12 scanning points is conducted using our method with the following parameters: for 12×12 scanning points, \(\sigma = 9 \times 10^{-6}\) and \(\lambda = 0.5\); for 6×6, \(\sigma = 9.5 \times 10^{-6}\) and \(\lambda = 0.5\); and for 4×4, \(\sigma = 10 \times 10^{-6}\) and \(\lambda = 0.5\). }
 	\label{nonconfocal-k}
 \end{figure*}

\begin{figure}[ht]
      \begin{center}			
      \includegraphics[width=1\linewidth]{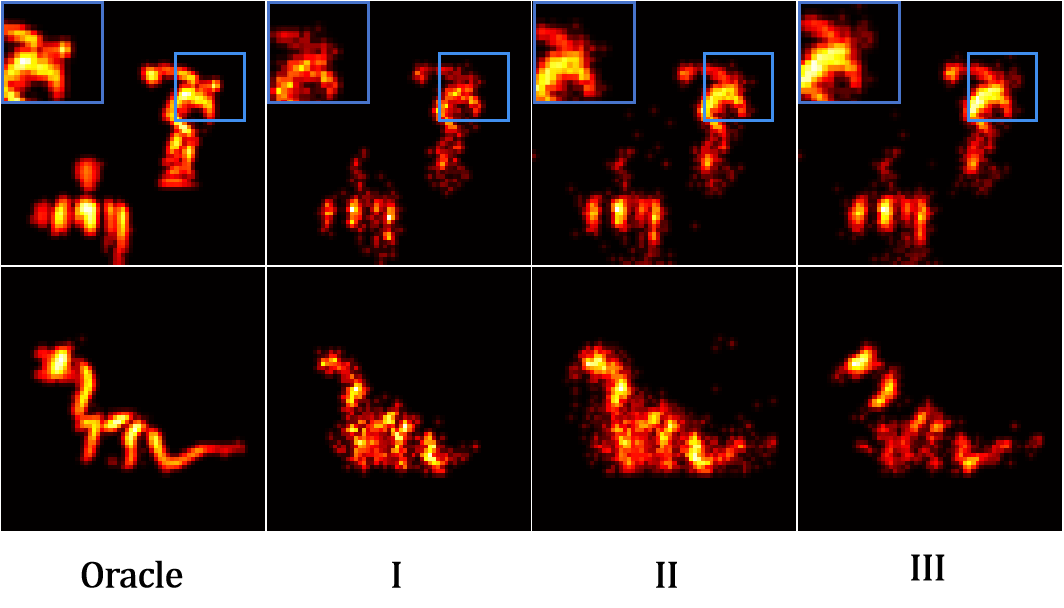}
	  \end{center}
	\caption{Comparison on `outdoor' (10 min) and `dragon' (60 min) scenes with 28$\times$28 and 16$\times$16 scanning points, respectively. The two oracles are generated using the PF method with 64$\times$64 measurements and exposure times of 50 min and 180 min. }
	\label{ablation-expe}
\end{figure}

Finally, we evaluate our SLCT method on a nonconfocal dataset, comparing it with non-confocal solvers SSCR and CNLOS. Additionally, using the midpoint approximation method from \cite{lindell2019wave}, we convert non-confocal data to confocal data and compare our method with LCT, DLCT, PF, and SPIRAL. As shown in Fig. \ref{nonconfocal-k}, our method consistently delivers the clearest and highest-resolution results across all three scanning setting. At 6×6 and 4×4 scanning points, methods other than SSCR, CNLOS, and ours exhibit noticeable degradation. Even at a 4×4 scanning points, our method achieves high-quality image reconstruction, outperforming other approaches.

\begin{table}[ht]
\footnotesize
\caption{Performance evaluation of our SLCT on `outdoor' and `dragon' scene, where `S.Rec.' denotes the sparse signal reconstruction module w.r.t. \cref{initialization of u}, `S.Res.' denotes the signal restoration module w.r.t. \cref{object-recon}, and `G.Con.' denotes the geometric constraints on depth image $D$ and albedo image $I$.}
\label{table:ablation}
\centering
\begin{minipage}{0.48\textwidth}
\begin{tabular}{@{\hskip .1pt}c|@{\hskip 3pt}c@{\hskip 3pt}c@{\hskip 3pt}c|c c|c c}
\toprule
\multirow{2}{*}{\textbf{  }} & \multicolumn{3}{c|}{\textbf{Methods}} & \multicolumn{2}{c|}{\textbf{outdoor}} & \multicolumn{2}{c}{\textbf{dragon}} \\
\cline{2-8}
 & \textbf{S.Rec.} & \textbf{S.Res.} & \textbf{G.Con.} &\textbf{PSNR$\uparrow$} & \textbf{SSIM$\uparrow$}&\textbf{PSNR$\uparrow$}  &\textbf{SSIM$\uparrow$}  \\
\midrule
\textbf{I}  & \ding{51}   & \ding{55} & \ding{55} & 19.738 & 0.800 & 19.998 &0.814 \\
\textbf{II} & \ding{51} & \ding{51} & \ding{55} & 21.260 & 0.794 & 20.363 &0.762\\
\midrule
\textbf{III}    & \ding{51} & \ding{51} & \ding{51} & \textbf{21.348} & \textbf{0.815} & \textbf{20.375}  &\textbf{0.817}\\
\bottomrule
\end{tabular}
 \end{minipage}
\end{table}

 \begin{table*}[t]
\footnotesize
\centering
\caption{Computational time complexity for a reconstruction region of $N \times N \times N$ and scanning points of $N \times N$.}
\label{table:computational complexity}
\begin{tabular}{@{\hskip 0pt}c|c|c|c|c|c|c|c}
\toprule
\multirow{2}{*} & \multicolumn{5}{c|}{\textbf{Regularization}} &  \multirow{2}{*}{Data term} & \multirow{2}{*}{In total} \\
\cline{2-6}
 & \textbf{Lasso (3D)} & \textbf{Dictionary (3D)} &  \textbf{Curvature (3D)} &\textbf{TV (2D)} & \textbf{Shape Operator (2D)} & &  \\
\midrule
\textbf{LCT\cite{o2018confocal}}    & /  & /  & /          & /          & /          & $O(N^3\log N)$        & $O(N^3\log N)$       \\
\textbf{DLCT\cite{young2020non}}    & /  & /  & /          & /          & /          & $O(N^3\log N)$        & $O(N^3\log N)$       \\
\textbf{PF\cite{liu2020phasor}}    & /  & /  & /          & /          & /          & $O(N^3\log N)$        & $O(N^3\log N)$       \\
\textbf{SPIRAL\cite{ye2021compressed}}    & $O(N^3)$  & /  & /          & /          & /          & $O(N^3\log N)$        & $O(N^3\log N)$       \\
\textbf{SSCR\cite{liu2023few}}   & $O(N^3)$  & $O(N^3)$  & /          & /          & /          & $O(N^5)$        & $O(N^5)$        \\
\textbf{CNLOS\cite{ding2024curvature}}  & /   & /       & $O(N^3\log N)$ & /            & /          & $O(N^3\log N)$ & $O(N^3\log N)$ \\
\midrule
\textbf{SLCT}   & $O(N^3)$  & /    & /     &$O(N^2 \log N)$     & $O(N^2 \log N)$   & $O(N^3\log N)$       & $O(N^3\log N)$       \\
\bottomrule
\end{tabular}
\end{table*}

\subsection{Complexity analysis}
\label{sec:conclusion}
We theoretically compare the computational and memory complexities of various algorithms. For a reconstruction region with discrete voxel dimensions of $N \times N \times N$, and a visible wall sampled at $N \times N$ points, the space complexity is $O(N^3)$, as the maximum data stored is the signal $\tau$ or the reconstructed image $u$. Table \ref{table:computational complexity} summarizes the time complexity of different methods. Except for SSCR, all methods have a computational time complexity of $O(N^3logN)$ for $N \times N$ scanning points.


\section{Conclusion and Discussion}
In this paper, we introduced a geometric-constrained NLOS imaging reconstruction model that applies constraints on surface normals using the shape operator. Accurate normal reconstruction is crucial to capture the details and contours of the object. Our model enhanced object reconstruction accuracy through  incorporating geometric constraints. Additionally, regularization on 2D depth and albedo maps significantly improved computational efficiency compared to volumetric albedo regularization. Experimental results demonstrated that the proposed method outperforms existing direct and iterative approaches in reconstruction quality under sparse illumination and low exposure conditions.

However, non-convex surfaces inherently introduce ambiguities, resulting in a many-to-one mapping between the actual geometry and its corresponding projection. This fundamentally challenges the uniqueness of the reconstruction process. A potential solution to this problem is to leverage the conformal geometry to eliminate self-overlapping \cite{lui2013shape,chen2024robust}, thereby improving the stability and reliability of the reconstruction. In future work, we aim to use conformal geometry to create a one-to-one mapping between the surface and two-dimensional depth and intensity maps, which can greatly facilitate higher-quality reconstructions for both traditional model-based and deep learning-based approaches.

\section*{Acknowledgments}
The work is supported by the National Natural Science Foundation of China (NSFC 12071345). Yuping Duan is the corresponding author.

\begin{IEEEbiography}[{\includegraphics[width=1in,height=1.25in,clip,keepaspectratio]{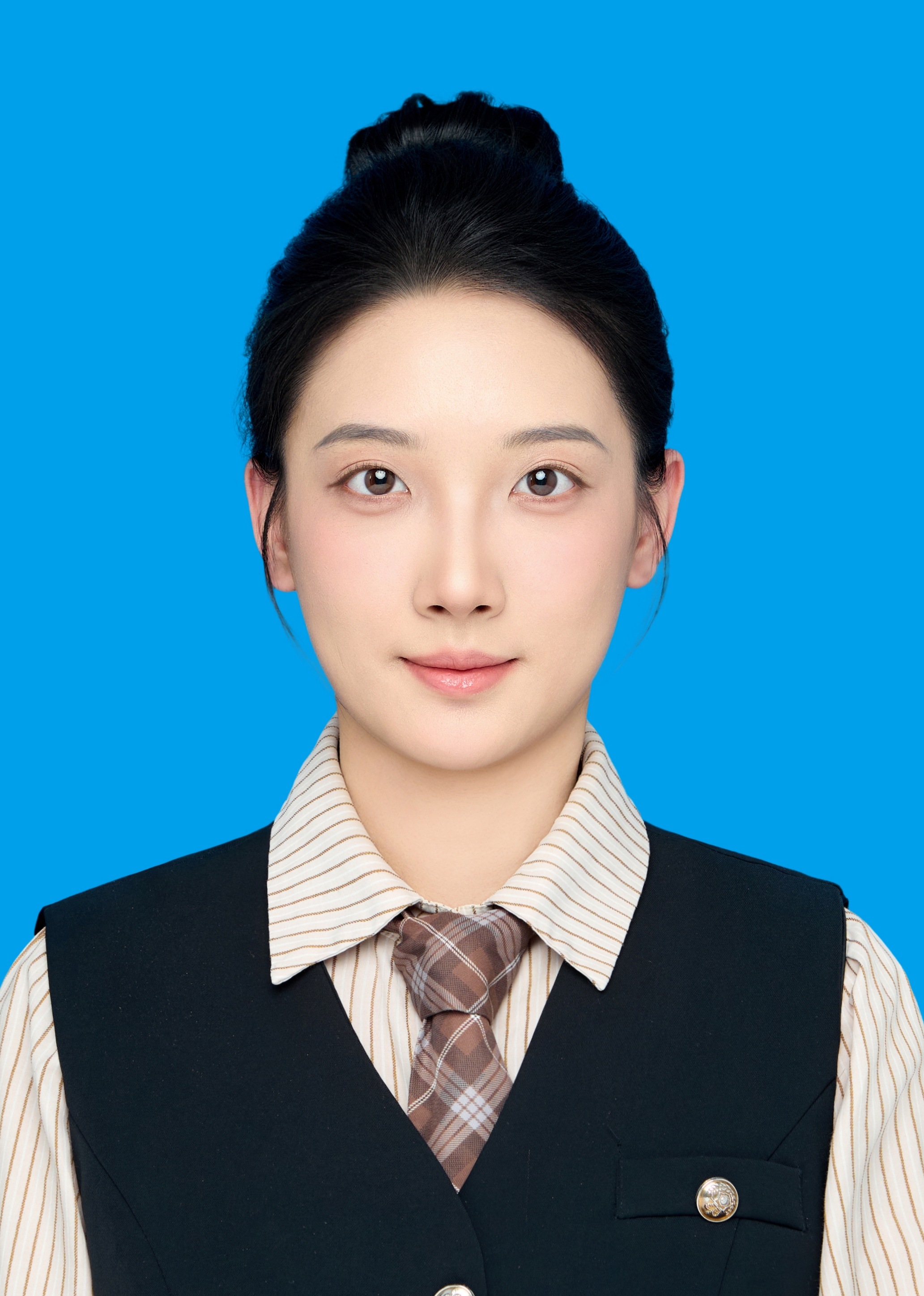}}]{Xueying Liu}
is currently a master's candidate at the Center for Applied Mathematics of Tianjin University. Her research direction is non-line-of-sight imaging.
\end{IEEEbiography}
\begin{IEEEbiography}[{\includegraphics[width=1in,height=1.25in,clip,keepaspectratio]{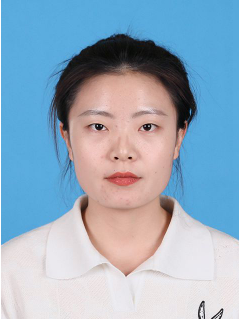}}]{Lianfang Wang}
 is currently pursuing a Ph.D. at Beijing Normal University. Her research focuses on deep learning-based methods applied to imaging fields, including computed tomography imaging and non-line-of-sight imaging.
\end{IEEEbiography}
\begin{IEEEbiography}[{\includegraphics[width=1in,height=1.25in,clip,keepaspectratio]{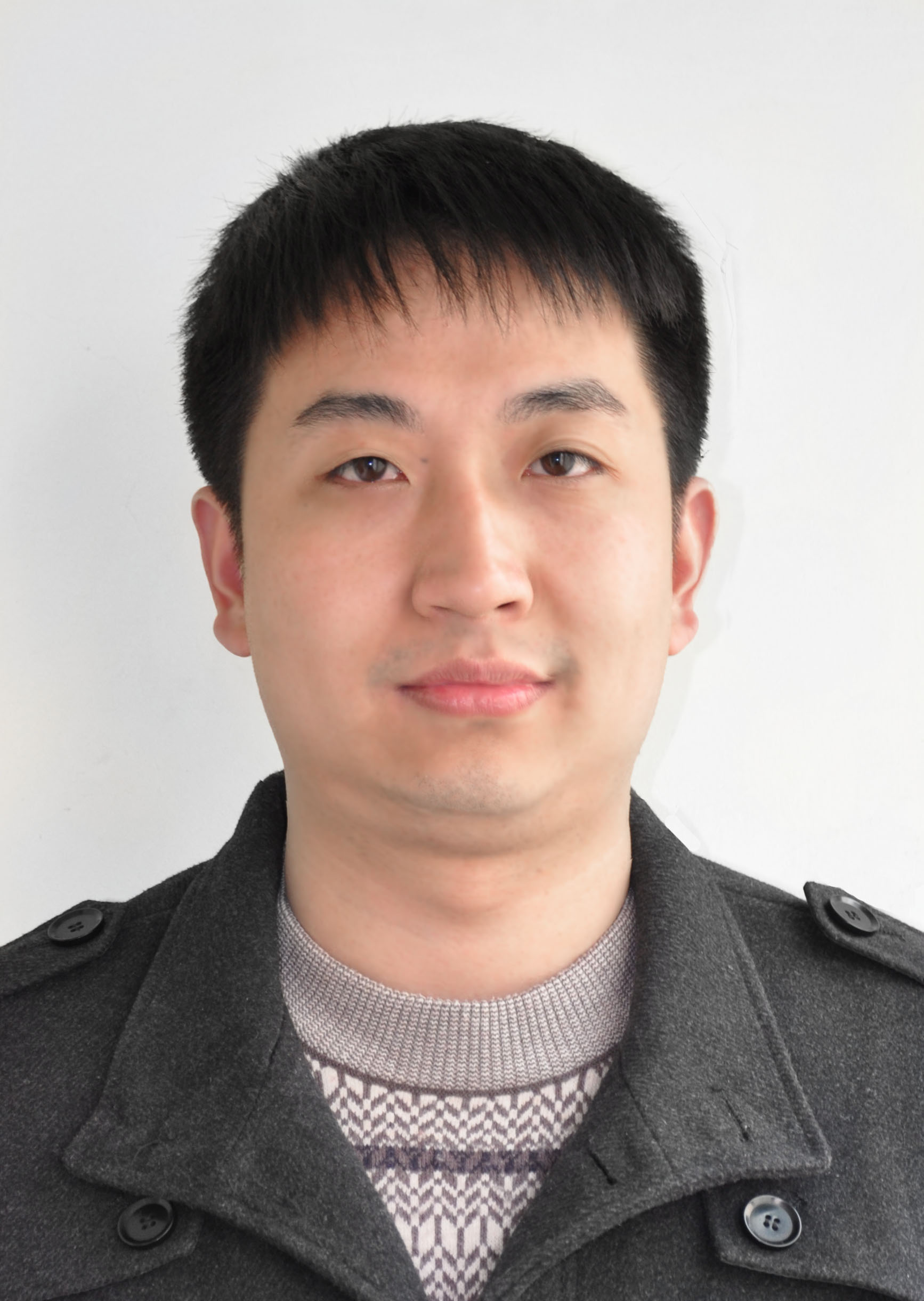}}]{Jun Liu}
 received the B.S. degree in mathematics from the Hunan Normal University, China in 2004. He received the M.S. and Ph.D. degrees in computational mathematics from the Beijing Normal University (BNU), China, in 2008 and 2011 respectively. He is currently an associate professor at BNU. His research interests include variational, optimal transport and deep learning based image processing algorithms and their applications.
\end{IEEEbiography}
\begin{IEEEbiography}[{\includegraphics[width=1in,height=1.25in,clip,keepaspectratio]{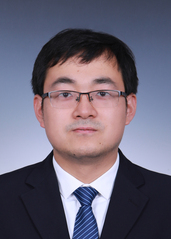}}]{Yong Wang}
received the Ph.D. degree from the
Institute of Physics, Chinese Academy of Sciences,
Beijing, China, in 2009. He is currently an Associate Professor with the School of Physics, Nankai University, Tianjin,
China. His research interests are theoretical and
computational physics on materials and devices.
\end{IEEEbiography}
\begin{IEEEbiography}[{\includegraphics[width=1in,height=1.25in,clip,keepaspectratio]{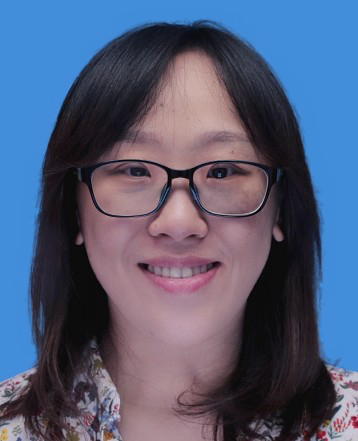}}]{Yuping Duan}
 is a full professor at the School of Mathematical Sciences of Beijing Normal University (BNU). Before joining BNU, she was a professor at Tianjin University in 2015 to 2023, and a research scientist at I2R, A*STAR in 2012 to 2015. She received her Ph.D. from Nanyang Technological University in 2012. Her research interests are image processing and computer vision, variational methods, and deep learning methods.
\end{IEEEbiography}




\end{document}